\documentclass[11pt]{article}

\usepackage{mathtools}
\usepackage{agenticlearning-xelatex}

\usepackage{amsmath}
\usepackage{amsthm}
\usepackage{enumitem}
\usepackage{multirow}
\usepackage{makecell}
\usepackage{mdframed}
\usepackage{wrapfig}

\begin{document}

% ---------------------------------------------------------------
% Set short title and author for headers
\shorttitle{A Beta-Bernoulli Calibrator for LLM Forecasting}
\shortauthor{Hui Dai \etal}

\title{Aligning LLMs with Human Uncertainty: \\A Beta-Bernoulli Calibrator for LLM Forecasting}

\author{
  Hui Dai$^{1,2}$, Ryan Teehan$^{1}$, Parsa Torabian$^{3}$, Mengye Ren$^{1}$ \\
  $^{1}$Agentic Learning AI Lab, New York University, $^{2}$The University of Chicago, $^{3}$Chronologies AI\\
  \texttt{\{hd2584, mengye\}@nyu.edu}\\
  \url{https://agenticlearning.ai/beta-bernoulli-calibrator}
}
\date{May 26, 2026}
\maketitle
\begin{abstract}
Probabilistic forecasting estimates the likelihood of uncertain future events. To improve LLM forecasting, existing methods typically learn from binary outcomes to output verbalized forecasts. However, while aggregated human forecasts contain rich information in both the crowd probability estimate and the degree of agreement among forecasters, how to utilize these signals remains underexplored. To address this, we propose the Beta-Bernoulli Calibrator (BBC), which converts an initial point estimate forecast from any model into a distribution over event likelihood, using supervision from both binary outcomes and human forecasts. BBC models event likelihood $p \sim \text{Beta}(\alpha, \beta)$ and outcome $y \sim \text{Bernoulli}(p)$, with the mean as the calibrated point forecast and the variance as the epistemic uncertainty. Our results show that BBC generally provides better calibrated and more accurate forecasts than both traditional post-hoc calibration methods and models fine-tuned specifically for forecasting, while remaining lightweight and having good generalization. We also show that the epistemic uncertainty captured by BBC is a more reliable predictor of forecasting error than verbalized confidence.
\end{abstract}

\section{Introduction}
\label{sec:intro}
Making predictions about the future is an integral part of everyday decision-making. Individuals check weather forecasts to adjust travel plans, companies calculate the odds of a product's success, and governments shape policy around economic and national security forecasts \citep{lahiri2013forecasting, tetlock2016superforecasting}. Given large language models' (LLMs) broad knowledge and reasoning capabilities, there is increasing interest in using LLMs for forecasting, typically by prompting the model to output a verbalized estimate of an event's likelihood \citep{karger2024forecastbench, zeng2025futurex, yang2026llm}. However, even state-of-the-art models struggle to outperform skilled human forecasters \citep{karger2024forecastbench}. 

To improve forecasting capabilities, prior work has investigated supervised fine-tuning via distillation on subsets where the model outperforms humans \citep{halawi2024approaching}, as well as reinforcement learning (RL) using signals from realized outcomes \citep{chandak2025scaling, turtel2026future}. However, these approaches are resource-intensive and typically cannot be applied to black-box models. Moreover, human forecasts contain rich information about human sentiment and uncertainty, as they capture both the aggregate estimate of an event's likelihood and the amount of consensus among the pool of forecasters. In spite of this, incorporating this information remains underexplored, and current methods do not capture the degree of consensus among the human forecasters. In this work, we ask: \textit{beyond eliciting verbalized probabilities, how can we calibrate model forecasts using supervision from both binary outcomes and human forecasts?}

\begin{figure*}[t]
  \begin{center}
    \centerline{\includegraphics[width=\textwidth]{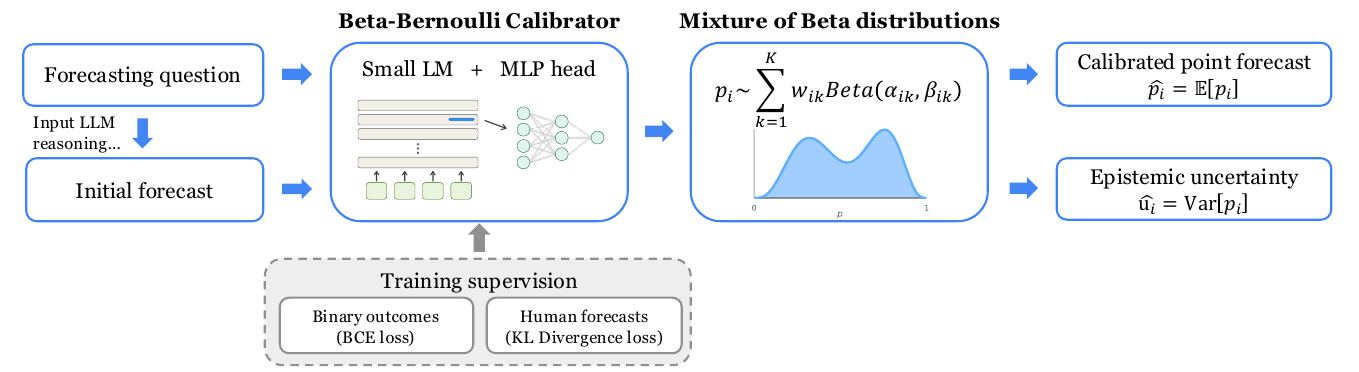}}
    \caption{
      Overview of the Beta-Bernoulli Calibrator (BBC). Given a forecasting question and an initial verbalized forecast from an input LLM, BBC outputs a mixture of Beta distributions over the event probability. BBC is itself a small language model with an MLP head that predicts the Beta parameters, trained using supervision from both binary outcomes and human forecasts. The mean of the predicted distribution serves as the calibrated point forecast and the variance as epistemic uncertainty.}
    \label{fig:pipeline}
  \end{center}
  \vskip -0.3in
\end{figure*}

As illustrated in Figure~\ref{fig:pipeline}, we propose a Beta-Bernoulli framework in which the event probability $p$ is modeled as a Beta distribution, $p \sim \text{Beta}(\alpha, \beta)$, and the observed outcome $y$ is a realization from a Bernoulli trial, $y \sim \text{Bernoulli}(p)$. In practice, we relax this single Beta to a mixture of $K$ Betas for added flexibility. Taking the forecasting question and an initial verbalized forecast as input, the calibrator outputs the Beta parameters. Notably, this framework is model-agnostic. It is implemented using a small, open-source language model (the calibrator) to refine the initial textual forecast provided by a separate input LLM. This allows us to calibrate any input LLM's beliefs without access to its internal representations, ensuring universal applicability and reducing training overhead. When learning from only binary outcomes, we show that the Beta–Bernoulli objective reduces to binary cross-entropy (BCE), a proper scoring rule that incentivizes truthful probability estimation. To incorporate signals beyond the binary outcome, we use human forecasts as distributional supervision for the Beta distribution. This enables the model to represent both the predicted event probability via the Beta mean, and epistemic uncertainty about that probability via the Beta variance.

We evaluate our framework on data from prediction platforms Metaculus and Polymarket. Compared to uncertainty estimation and post-hoc calibration methods, our Beta-Bernoulli Calibrator (BBC) generally provides better-calibrated forecasts with stronger discrimination performance. We find that utilizing human forecasts as auxiliary supervision consistently improves discrimination compared to training only on binary outcomes. Moreover, this lightweight post-hoc adjustment even outperforms models that are fine-tuned specifically for forecasting, and provides further improvements when applied to them. In addition, we validate that BBC's epistemic uncertainty is a strong predictor of forecasting errors, while verbalized confidence is a noisier signal. Finally, we test our calibrator's generalization on the external Kalshi dataset, and observe consistent performance gain. Therefore, we present the following contributions:
\begin{itemize}[leftmargin=*]
    \item \textbf{Beta-Bernoulli Calibrator.} We propose BBC, a lightweight, model-agnostic calibrator that converts an initial probability forecast into a distribution over event likelihood. This effectively captures both aleatoric and epistemic uncertainty in event forecasting.
    
    \item \textbf{Humans as Distributional Supervision.} Human forecasts are a rich source of data that have so far been underutilized. In addition to providing an aggregated estimate of an event's probability, the degree of consensus among the forecasts provides information about human sentiment and uncertainty. We use these forecasts as distributional supervision, which significantly improves AUC and allows us to go beyond only learning from binary outcomes.
\end{itemize}

\section{Related work}

\paragraph{Traditional calibration and evidential methods.}
Earlier work on uncertainty calibration mainly focused on post-hoc calibration of classifier outputs \citep{degroot1983comparison,niculescu2005predicting}. Parametric methods such as Platt scaling \citep{platt1999probabilistic} and temperature scaling \citep{guo2017calibration} learn global parameters to rescale prediction scores across all samples. For nonparametric methods, histogram binning \citep{zadrozny2001obtaining} uses the empirical outcome frequencies in bins as calibrated scores, and isotonic regression \citep{zadrozny2002transforming} learns a monotonic piecewise constant function to transform uncalibrated scores. Post-hoc calibration relates to BBC’s role as a calibrator, while \emph{Evidential Deep Learning} (EDL) relates to its probabilistic output parameterization. EDL models a Dirichlet over categorical probability predictions, with the Beta as the binary special case \citep{sensoy2018evidential, charpentier2020posterior}. However, our framework differs in two ways: (i) rather than collecting evidence directly from task inputs (like a cat image) in an end-to-end classifier, BBC is a stagewise calibrator that adjusts on top of another model's natural language output, benefiting from its reasoning capability; (ii) rather than learning only from deterministic class labels, we introduce learning from human forecast distributions, which provides additional supervision and helps address the identifiability issue discussed in Section~\ref{sec:objective_functions}.

\paragraph{Uncertainty estimation and calibration in LLMs.}
In the context of LLMs, the focus shifts from calibrating classifier scores to estimating the reliability of natural language generations \citep{shorinwa2025survey}. Most work has concentrated on tasks such as mathematics (e.g. GSM8K \citep{cobbe2021training}) and reasoning (e.g. HotpotQA \citep{yang2018hotpotqa}).
% , and reading comprehension (e.g.TriviaQA \citep{joshi2017triviaqa}). 
In these settings, the model generates an answer and the elicited confidence score should reflect the probability that the answer is correct. We survey uncertainty estimation and calibration in LLMs into training-free and training-based categories:

\noindent \textit{Training-free methods} extract uncertainty estimates without modifying model weights.
In black-box settings, \textit{verbalized uncertainty} can be obtained simply by prompting the model to state its confidence after providing an answer \citep{tian2023just}. However, such self-reported confidence is found to be overconfident \citep{xiong2024can, mei2026reasoning, kirichenko2025abstentionbench}. White-box methods instead leverage internal model features. These are primarily \textit{logit-based}, estimating uncertainty through the entropy of output-token probabilities  \citep{ling2024uncertainty, fadeeva2024fact}. Another popular approach is $P(\text{True})$, where the model is prompted to assess whether its own answer is ``True'' or ``False,'' and the probability of getting the ``True'' token is interpreted as its confidence score \citep{kadavath2022language}. Finally, sampling-based ensembles (such as majority vote or taking average) can be applied to both verbalized and logit-based methods to further improve calibration \citep{zhang2024luq,jiang2023calibrating,xiong2024can}.

\noindent \textit{Training-based methods} learn calibrated confidence predictors or elicit better-calibrated uncertainty through training.
Although some studies find that verbalized confidence or simple token-based signals can be well-calibrated \citep{tian2023just,kadavath2022language}, other work shows they underperform training-based approaches \citep{kapoor2024large}. 
A primary direction probes internal representations, as hidden layers have been shown to encode information regarding truthfulness and potential error patterns \citep{orgad2025llms}. These methods train \textbf{probing classifiers} on top of LLM hidden states to predict answer correctness \citep{kadavath2022language, azaria2023internal, kapoor2024large, zhang2025reasoning}. Beyond add-on probes, another line \textbf{fine-tunes} the LLM itself to express calibrated uncertainty in natural language. For example, \citet{lin2022teaching} fine-tune GPT-3 using the model's empirical accuracy across different question types as a proxy for ground truth confidence. More recently, work has explored the use of proper scoring rules as fine-tuning objectives \citep{li2025conftuner} or incorporating calibration-aware reward functions in RL to incentivize honest confidence reporting \citep{xu2024sayself, damani2026beyond}. 

\paragraph{LLMs in forecasting.}
The uncertainty work reviewed above treats uncertainty as confidence in an answer's correctness. While the tasks are useful for measuring model performance, they primarily address epistemic uncertainty, which arises from a model's lack of knowledge and is, in principle, reducible \citep{kendall2017uncertainties}. That is, tasks such as mathematical problem solving do not involve inherent randomness, and a perfect system should always produce the correct answer with confidence 1.0. In forecasting, by contrast, uncertainty estimation is not merely a diagnostic measure of confidence, but the primary output of interest for predicting future events. Real-world events such as market fluctuations or weather patterns possess aleatoric uncertainty, or irreducible randomness inherent to the event itself. Current work typically prompts LLMs to provide verbalized probability estimates \citep{karger2024forecastbench, zeng2025futurex, yang2026llm}, which are often overconfident \citep{schoenegger2024wisdom, halawi2024approaching, nel2025large}. To improve the forecasts, \citet{alur2025aia} apply ensembling and traditional post-hoc calibration, \citet{murphy2026agentic} combine an agentic search loop with hierarchical Platt scaling, and \citet{halawi2024approaching} fine-tune GPT-4 on subsets where model outperforms human crowd. Recent efforts explore RL, using Brier score and accuracy as reward signals for open-ended forecasting \citep{chandak2025scaling}, and binary cross-entropy for binary prediction tasks \citep{turtel2026future}. While prior work targets verbalized point forecasts, our work is
% , to the best of our knowledge, 
the first to utilize human forecast signals to model the distribution over event probabilities. Moreover, as a post-hoc calibrator, our method is complementary to these methods: it can be applied on top of them to further improve their forecasts.

\section{Preliminaries}
\subsection{Problem setup}
We study the task of probabilistic forecasting for binary events \citep{lahiri2013forecasting}. Let $D = \{(x_i, y_i, \mathbf{q}_i)\}_{i=1}^N$ be a dataset of $N$ binary forecasting questions, where $x_i$ is the textual description of event $i$ (e.g., a question and its resolution criteria), and $y_i \in \{0,1\}$ denotes the binary outcome. In addition, we observe $k_i$ human forecasts for each event
$\mathbf{q}_i = \{q_{i1}, \dots, q_{ik_i}\}$, where $q_{ij}\in[0,1]$ is the probability estimate provided by forecaster $j$. We assume $p_i^\star = P(y_i=1 \mid x_i)$ is the unobservable ground-truth event probability. Our goal is to learn a model $f_\theta$ that takes in $x_i$ and outputs a probability forecast $\hat{p}_i \in [0,1]$, such that $\hat{p}_i \approx p^*_i$. Note that prior work typically extracts $\hat{p}_i$ from verbalized output, e.g. set $\hat{p}=0.2$ if the model outputs ``I estimate a 20\% chance''. In contrast, our framework models $p_i$ as a distribution and later reports the mean as the point estimate $\hat{p_i} = \mathbb{E}[p_i]$ (see Section \ref{sec:bbc}).

\subsection{Evaluation metrics}
We use the following forecasting metrics to evaluate $\hat{p_i}$: (i) \textbf{Brier score} \citep{brier1950verification}, the mean squared error between the predicted probabilities $\hat{p}$ and the binary outcomes $y_i$; (ii) \textbf{Accuracy}, the fraction of correct predictions after thresholding $\hat{p}_i$ at $0.5$; (iii) \textbf{AUC} \citep{bradley1997use}, measuring threshold-free discrimination performance; and (iv) \textbf{Expected Calibration Error (ECE)} \citep{naeini2015obtaining}, which measures how uncalibrated a model is by taking the expectation of the absolute difference between the model prediction and the empirical event occurrences, computed from the set of events with similar predictions.\footnote{For additional details in evaluation metrics, see Appendix \ref{app:metrics}.}

\section{Beta-Bernoulli Calibrator}
\label{sec:bbc}
\subsection{Overview} 
\label{sec:bbc_overview}
Figure~\ref{fig:pipeline} summarizes our framework. The outcome $y_i$ is modeled as a Beta-Bernoulli process: first an event probability $p_i$ is drawn from a Beta distribution, and then $y_i$ is drawn from $\text{Bernoulli}(p_i)$. That is,
$p_i \sim \text{Beta}(\alpha_i, \beta_i)$ and $y_i \sim \text{Bernoulli}(p_i).$
Our model $f_\theta$ maps the input $x_i$ to the parameters of this Beta distribution, i.e., 
$f_\theta(x_i) = (\hat\alpha_i, \hat\beta_i), \text{where } \hat\alpha_i, \hat\beta_i > 0.$ The mean of the predicted distribution $\text{Beta}(\hat\alpha_i, \hat\beta_i)$ serves as the calibrated point estimate, and the variance as the epistemic uncertainty about the latent event probability $p_i$:
$
\hat{p}_i = \mathbb{E}[p_i] = \frac{\hat\alpha_i}{\hat\alpha_i+\hat\beta_i}, 
\hat{u}_i=\text{Var}[p_i] = \frac{\hat\alpha_i \hat\beta_i}{(\hat\alpha_i+\hat\beta_i)^2(\hat\alpha_i+\hat\beta_i+1)}.
$ Note that this epistemic uncertainty is the calibrator's learned estimate of uncertainty about $p_i$, distinct from the input LLM's internal confidence in its own forecast (which the calibrator does not have access to).

\subsection{Model architecture and input} Since the input $x_i$ is in natural language, we parameterize $f_\theta$ as a language model encoder followed by an MLP head that outputs Beta parameter values. We include an initial forecast $\hat{p_i}^{\text{init}}$ as part of the input, and train $f_\theta$ to act as a post-hoc calibrator that refines this initial belief. While our framework imposes no constraints on the source of the initial belief, for our experiments we derive $\hat{p}_i^{\text{init}}$ by prompting a separate LLM (input LLM) for verbalized probability. This follows the prior work, and offers both simplicity and broad applicability. Therefore, our input takes the form:
$x_i = \text{``Question: } \{text_i\}; \text{Initial forecast: } \{\hat{p}^{\text{init}}_i\}\text{''}.$

Importantly, note that our calibrator $f_\theta$ is model-agnostic in terms of the input LLM. This allows us to calibrate forecasts from any black-box models, thus we can leverage strong proprietary LLMs without fine-tuning them. Furthermore, because the task of calibration is distinct from the heavy reasoning required for the initial forecast, $f_\theta$ can be significantly smaller than the input LLM. We show in Section \ref{sec:ablation_model_size} that a small 1-billion parameter language model is sufficient to effectively calibrate initial forecasts from larger models, making our method computationally efficient.

\subsection{Objective functions}
\label{sec:objective_functions}
Given $D=\{(x_i,y_i,\mathbf{q}_i)\}_{i=1}^N$, we train the calibrator $f_\theta$ using supervision from both binary outcomes $y_i\in\{0,1\}$ and human forecasts $\mathbf{q}_i=\{q_{i1},\dots,q_{ik_i}\}$, thus the overall training objective combines both signals:
$
\mathcal{L}_{\text{total}} = \sum_{i=1}^N \mathcal{L}_{\text{binary},i} + \sum_{i=1}^N \mathcal{L}_{\text{human},i}.
$\footnote{We find performance to be relatively robust across a broad range of weightings between $\mathcal{L}_{\text{binary}}$ and $\mathcal{L}_{\text{human}}$ in Appendix~\ref{app:ablation_loss_coef}.}

\paragraph{Learning from binary outcomes.} We show that $\mathcal{L}_{\text{binary}}$ is equivalent to the Binary Cross-Entropy (BCE) loss. By marginalizing out $p$, the marginal likelihood of $y_i$ is:
\begin{align*}
P(y_i | \alpha_i,\beta_i) &= \int_0^1 P(y_i|p_i)\,P(p_i|\alpha_i,\beta_i) \mathop{}\mathrm{d} p_i =\int_0^1 p_i^{y_i}(1-p_i)^{1-y_i} \frac{1}{B(\alpha_i,\beta_i)}\,p_i^{\alpha_i-1}(1-p_i)^{\beta_i-1} \mathop{}\mathrm{d} p_i \\
&=\frac{1}{B(\alpha_i,\beta_i)}\int_0^1 p_i^{\alpha_i+y_i-1}(1-p_i)^{\beta_i+1-y_i-1} \mathop{}\mathrm{d} p_i =\frac{B(\alpha_i+y_i,\;\beta_i+1-y_i)}{B(\alpha_i,\beta_i)}.
\end{align*}

This equals $\frac{\alpha_i}{\alpha_i+\beta_i}$ when $y_i=1$, and $\frac{\beta_i}{\alpha_i+\beta_i}$ when $y_i=0$. Applying this to the predicted parameters, with $\hat{p}_i =  \frac{\hat\alpha_i}{\hat\alpha_i+\hat\beta_i}$, the Beta-Bernoulli loss reduces exactly to the BCE loss with respect to the mean $\hat{p}_i$:
$
    \mathcal{L}_{\text{binary},i} = -\log P(y_i | \alpha_i, \beta_i) = -y_i \log(\hat p_i) - (1-y_i)\log(1-\hat p_i). 
$
BCE is a strictly proper scoring rule \citep{gneiting2007strictly}, which incentivizes learning true probability $p^*_i$ as it is minimized if and only if $\hat{p}_i = p^*_i$. 

\paragraph{Learning from human forecasts.}
\looseness=-1
Learning from only binary outcomes is insufficient to capture a meaningful distribution under limited data.\footnote{In theory, infinite samples from the latent probability distribution would identify the ground-truth Beta parameters with BCE loss. However, in practice each event resolves to only one binary outcome, making the distribution shape hard to learn without additional signals.} There is an identifiability problem where the loss is invariant to the scale of Beta parameters (see a toy experiment validating this in Appendix~\ref{app:toy}). For example, $\text{Beta}(30,20)$ and $\text{Beta}(3,2)$ have the same mean $\hat{p}_i = 0.6$ and thus the same BCE loss, while they have different shapes and the latter is flatter with higher epistemic uncertainty. Moreover, as we see human forecasts as noisy samples from the true distribution over $p$, they not only provide the missing signal about distribution shape but also introduce supervision beyond the single binary outcome, enriching the information available per question. To learn from them, we can simply match our predicted Beta distributions with human forecast histograms via a Kullback-Leibler (KL) divergence objective. Let $\mathbf{h}_i$ be the normalized human forecast histogram over $B$ bins. We minimize $\mathcal{L}_{\text{human},i} = \text{KL} (\mathbf{h}_i || \text{Beta}(\alpha_i,\beta_i)).$

\subsection{Relaxing the constraint by mixture of Beta}
The above models $p$ as a single Beta distribution, which can be limited when the true underlying belief is multi-modal (e.g., when opinions are polarized). To further relax the prior family, in our experiments, we model $p$ as a mixture of $K$ Beta distributions. Therefore, the output dimension expands to $K$ pairs of $(\alpha, \beta)$ with corresponding weights. Concretely, with $\alpha_{ik},\beta_{ik}>0,w_{ik}\ge 0,\sum_{k=1}^K w_{ik}=1$,
$f_\theta^{\text{mixture}}(x_i) = \{(\alpha_{ik},\beta_{ik},w_{ik})\}_{k=1}^K$, and $p_i \sim \sum_{k=1}^K w_{ik} \text{Beta}(\alpha_{ik},\beta_{ik}).
$
During training, we optimize $\mathcal{L}_{\text{binary},i}$ in terms of the mixture mean $\hat p_i = \mathbb{E}[p_i]
= \sum_{k=1}^K \hat{w}_{ik}\,\frac{\hat{\alpha}_{ik}}{\hat{\alpha}_{ik}+\hat{\beta}_{ik}}$, and match the mixture distribution to human forecast histogram for $\mathcal{L}_{\text{human},i}$.

\section{Experiments}

\subsection{Dataset}
\label{sec:dataset}
We collect binary questions from the forecasting platforms Metaculus and Polymarket. To ensure data quality and sufficient crowd signal, we filter out low-volume questions and exclude domains that are hard to model, such as sports, weather, and cryptocurrency.\footnote{See Appendix \ref{app:dataset} for more details in data preprocessing and data distribution.} In total, this results in 11,355 resolved questions. As shown in Table \ref{tab:data_stat}, we split the data temporally: 7,824 training questions resolved before April 2025, 1,917 validation questions resolved between April and July 2025, and 1,614 test questions resolved between August 2025 and January 2026. This testing phase occurs entirely beyond the knowledge cutoff of all LLMs we tested, preventing data leakage. 

While both platforms provide human forecast information, the nature of this information is different. On Metaculus, a user $j$ can submit a forecast probability $q_{ij} \in[0,1]$ for event $i$, and we can directly get a 100-bin forecast histogram $\mathbf{h}_i$ from the API. On Polymarket, users trade yes/no contracts, whose prices can be interpreted as the market's consensus probability of the event. In that case, we can only construct a proxy histogram $\mathbf{h}_i$ by binning the market prices over a time window (between the market open time and close time), capturing the temporal volatility. As a result, the Metaculus histogram reflects explicit crowd agreement across different forecasters, while the Polymarket histogram reflects agreement of aggregate market beliefs over time. Nevertheless, they both provide informative human signals about the uncertainty in the underlying event probability.

\subsection{Experimental setup}
\label{sec:experimental_setup}
\paragraph{Training details.}
We choose Llama-3.2-1B \citep{grattafiori2024llama} as the base of our Beta-Bernoulli Calibrator $f_{\theta}$, and model $p$ as a mixture of $K=5$ Beta distributions for flexibility (see Appendix~\ref{app:ablation_K} for an ablation on $K$). The calibrator $f_{\theta}$ encodes input $x_i$ (question and initial forecast), takes the second-to-last-layer hidden state at the final non-padding token as the sequence representation, and maps it through a two-layer feed-forward head to predict the Beta parameters. To prevent overly extreme predictions, we constrain $\alpha_{ik},\beta_{ik} > 1$.

We test our framework using initial forecasts from 7 LLMs. All initial forecasts are generated using greedy decoding (temperature = 0). During training, the final MLP head is trained and the base LLM is fine-tuned with Low-Rank Adapters (LoRA) \citep{hu2022lora}. We sweep hyperparameters of LoRA rank $r \in \{128, 256\}$ (with LoRA scaling $\alpha = r$) and learning rates $\lambda \in \{1\text{e-}6, 5\text{e-}6\}$ over 3 random seeds, training for 15 epochs and selecting the best models with validation Brier score. Final results are reported as the average and standard deviation of the top-5 models on the test set. For all ablation studies, we report results averaged over 3 random seeds with fixed $r=256$ and $\lambda=1\text{e-}6$. All experiments are run on a single NVIDIA L40S or H200 GPU.

\paragraph{Baseline methods.}
We evaluate the Beta-Bernoulli Calibrator in two configurations: trained only on binary outcome labels (\textbf{BBC, binary only}) and trained on both binary outcomes and human forecasts (\textbf{BBC, binary + human}). This allows us to see the effect of learning from human uncertainty. We compare against both uncertainty estimation/calibration methods and models fine-tuned specifically for forecasting:
\begin{itemize}[leftmargin=*]
    \item \textbf{Verbalized:} We directly prompt the LLM to state probabilistic forecasts. These estimates serve as initial forecasts for the calibration methods (including ours, Platt Scaling and Isotonic Regression), whose goal is to improve this baseline. The prompt can be found in Appendix \ref{app:prompts}. 
    \item \textbf{Ensemble:} We prompt the LLM $n$ times and average the forecasts as $\hat p$.
    \item $\textbf{P(}\mathrm{\textbf{True)}}$~\citep{kadavath2022language}: A logit-based uncertainty estimation method that prompts the model to explicitly answer ``Yes'' or ``No'', and derives $\hat p = \frac{P(\text{Yes})}{P(\text{Yes}) + P(\text{No})}$. This is only feasible in white-box models.
    \item \textbf{Platt Scaling}~\citep{platt1999probabilistic}: A parametric calibration method that models $\hat p = \sigma(A \hat p ^{\text{init}} + B)$. The parameters $A$ and $B$ are learned by minimizing the negative log-likelihood on validation set.
    % $$
    % A, B
    % = \arg\min_{A,B}- \sum_{i=1}^N\bigl(y_i \log \hat p_i + (1 - y_i)\log(1 - \hat p_i)\bigr).
    % $$
    \item \textbf{Isotonic Regression}~\citep{zadrozny2002transforming}: A non-parametric calibration method that learns a piecewise constant function by minimizing squared error under an order constraint.
    \item \textbf{OpenForecaster-8B}~\citep{chandak2025scaling}\footnote{\url{https://huggingface.co/nikhilchandak/OpenForecaster-8B}}: A Qwen3-8B \citep{yang2025qwen3} model fine-tuned with RL, using accuracy and Brier score as rewards. In addition to the Metaculus binary questions, the model uses 52K synthetically generated open-ended forecasting questions from news articles. Their training data cutoff is April 2025, consistent with the temporal split of ours.
    \item \textbf{Future-as-a-label-32B}~\citep{turtel2026future}\footnote{\url{https://huggingface.co/LightningRodLabs/future-as-label-paper-step160}}: A Qwen3-32B \citep{yang2025qwen3} model fine-tuned with RL, using BCE as reward. The training data consists 5,120 binary questions generated from news articles, with a cutoff date of January 30, 2025.
    
\end{itemize}

\subsection{Results}
\label{sec:results}

% \begin{table*}[!htbp]
\begin{table*}[!t]
  \centering
  \caption{Test performance across input LLMs and baseline methods. Best results are bolded, and second-best results are underlined. KL is the KL divergence between the predicted distribution and the human forecast distribution on the test set.}
  \label{tab:results}
  \footnotesize
  \setlength{\tabcolsep}{3pt}
  \renewcommand{\arraystretch}{1}
  \begin{tabular}{@{}l rr rr rr rr rr@{}}
    \toprule
    \tableheader &
    \multicolumn{2}{c}{\tableheader Brier$\downarrow$} &
    \multicolumn{2}{c}{\tableheader Accuracy$\uparrow$} &
    \multicolumn{2}{c}{\tableheader AUC$\uparrow$} &
    \multicolumn{2}{c}{\tableheader ECE$\downarrow$} &
    \multicolumn{2}{c}{\tableheader KL$\downarrow$} \\
    \cmidrule(lr){2-3}\cmidrule(lr){4-5}\cmidrule(lr){6-7}\cmidrule(lr){8-9}\cmidrule(lr){10-11}
    \tableheader Input LLM / Method &
    \multicolumn{1}{c}{\tableheader mean} & \multicolumn{1}{c}{\tableheader std} &
    \multicolumn{1}{c}{\tableheader mean} & \multicolumn{1}{c}{\tableheader std} &
    \multicolumn{1}{c}{\tableheader mean} & \multicolumn{1}{c}{\tableheader std} &
    \multicolumn{1}{c}{\tableheader mean} & \multicolumn{1}{c}{\tableheader std} &
    \multicolumn{1}{c}{\tableheader mean} & \multicolumn{1}{c}{\tableheader std} \\
    \midrule

    \rowwhite \textbf{Human Baseline} & 0.061 &  & 0.923 &  & 0.958 &  & 0.055 &  &  &  \\
    \midrule

    \textbf{Claude-Sonnet-4} \\
    \rowwhite \hspace*{1.2em}Verbalized            & 0.146 &  & 0.799 &  & 0.723 &  & 0.104 &  &  &  \\
    \rowwhite \hspace*{1.2em}Ensemble ($n=3$)      & 0.143 &  & 0.800 &  & \secondbest{0.736} &  & 0.100 &  &  &  \\
    \rowwhite \hspace*{1.2em}Platt Scaling         & 0.129 &  & 0.827 &  & 0.723 &  & \secondbest{0.034} &  &  &  \\
    \rowwhite \hspace*{1.2em}Isotonic Regression   & 0.129 &  & 0.832 &  & 0.724 &  & 0.038 &  &  &  \\
    \rowwhite \hspace*{1.2em}BBC (binary only)     & \secondbest{0.128} & (0.001) & \secondbest{0.833} & (0.002) & 0.732 & (0.003) & 0.036 & (0.011) & 9.004 & (0.266) \\
    \rowblue  \hspace*{1.2em}BBC (binary+human)    & \best{0.125} & (0.002) & \best{0.837} & (0.004) & \best{0.742} & (0.007) & \best{0.027} & (0.006) & \best{8.775} & (0.319) \\
    \midrule

    \textbf{Llama-3.3-70B-Instruct} \\
    \rowwhite \hspace*{1.2em}Verbalized            & 0.157 &  & 0.777 &  & 0.655 &  & 0.119 &  &  &  \\
    \rowwhite \hspace*{1.2em}Ensemble ($n=10$)     & 0.151 &  & 0.782 &  & 0.669 &  & 0.099 &  &  &  \\
    \rowwhite \hspace*{1.2em}Platt Scaling         & 0.139 &  & \best{0.829} &  & 0.655 &  & 0.051 &  &  &  \\
    \rowwhite \hspace*{1.2em}Isotonic Regression   & \secondbest{0.138} &  & \secondbest{0.822} &  & 0.654 &  & \best{0.043} &  &  &  \\
    \rowwhite \hspace*{1.2em}$P(\mathrm{True})$    & 0.265 &  & 0.726 &  & 0.656 &  & 0.265 &  &  &  \\
    \rowwhite \hspace*{1.2em}BBC (binary only)     & \secondbest{0.138} & (0.003) & 0.816 & (0.011) & \secondbest{0.671} & (0.010) & 0.054 & (0.012) & 11.564 & (0.172) \\
    \rowblue  \hspace*{1.2em}BBC (binary+human)    & \best{0.135} & (0.002) & \best{0.829} & (0.003) & \best{0.679} & (0.006) & \secondbest{0.045} & (0.012) & \best{9.526} & (0.507) \\
    \midrule

    \textbf{Qwen3-32B} \\
    \rowwhite \hspace*{1.2em}Verbalized            & 0.158 &  & 0.796 &  & 0.661 &  & 0.111 &  &  &  \\
    \rowwhite \hspace*{1.2em}Ensemble ($n=10$)     & 0.143 &  & 0.813 &  & \best{0.700} &  & 0.097 &  &  &  \\
    \rowwhite \hspace*{1.2em}Platt Scaling         & 0.142 &  & 0.827 &  & 0.661 &  & 0.079 &  &  &  \\
    \rowwhite \hspace*{1.2em}Isotonic Regression   & 0.137 &  & 0.823 &  & 0.662 &  & \best{0.040} &  &  &  \\
    \rowwhite \hspace*{1.2em}$P(\mathrm{True})$    & 0.232 &  & 0.761 &  & 0.664 &  & 0.230 &  &  &  \\
    \rowwhite \hspace*{1.2em}Future-as-a-label-32B & 0.137 &  & 0.829 &  & 0.677 &  & 0.046 &  &  &  \\
    \rowwhite \hspace*{1.2em}BBC (binary only)     & \secondbest{0.135} & (0.005) & \secondbest{0.832} & (0.012) & 0.684 & (0.009) & \secondbest{0.044} & (0.015) & 11.085 & (0.453) \\
    \rowblue  \hspace*{1.2em}BBC (binary+human)    & \best{0.133} & (0.002) & \best{0.833} & (0.004) & \secondbest{0.686} & (0.004) & 0.046 & (0.014) & \best{9.402} & (0.405) \\
    \midrule

    \textbf{Qwen3-8B} \\
    \rowwhite \hspace*{1.2em}Verbalized            & 0.185 &  & 0.750 &  & 0.633 &  & 0.164 &  &  &  \\
    \rowwhite \hspace*{1.2em}Ensemble ($n=10$)     & 0.169 &  & 0.755 &  & 0.661 &  & 0.148 &  &  &  \\
    \rowwhite \hspace*{1.2em}Platt Scaling         & 0.151 &  & 0.823 &  & 0.633 &  & 0.094 &  &  &  \\
    \rowwhite \hspace*{1.2em}Isotonic Regression   & 0.141 &  & 0.818 &  & 0.638 &  & 0.054 &  &  &  \\
    \rowwhite \hspace*{1.2em}$P(\mathrm{True})$    & 0.222 &  & 0.772 &  & 0.619 &  & 0.220 &  &  &  \\
    \rowwhite \hspace*{1.2em}OpenForecaster-8B    & 0.157 &  & 0.794 &  & \secondbest{0.663} &  & 0.084 &  &  &  \\
    \rowwhite \hspace*{1.2em}BBC (binary only)     & \secondbest{0.138} & (0.001) & \secondbest{0.824} & (0.008) & 0.662 & (0.008) & \best{0.044} & (0.008) & 11.550 & (0.390) \\
    \rowblue  \hspace*{1.2em}BBC (binary+human)    & \best{0.137} & (0.004) & \best{0.828} & (0.004) & \best{0.673} & (0.016) & \secondbest{0.050} & (0.019) & \best{9.730} & (0.231) \\
    \bottomrule
  \end{tabular}
  \vspace{0.2in}
\end{table*}

\begin{figure}[h]
  
  \begin{center}
    \centerline{\includegraphics[width=0.65\columnwidth]{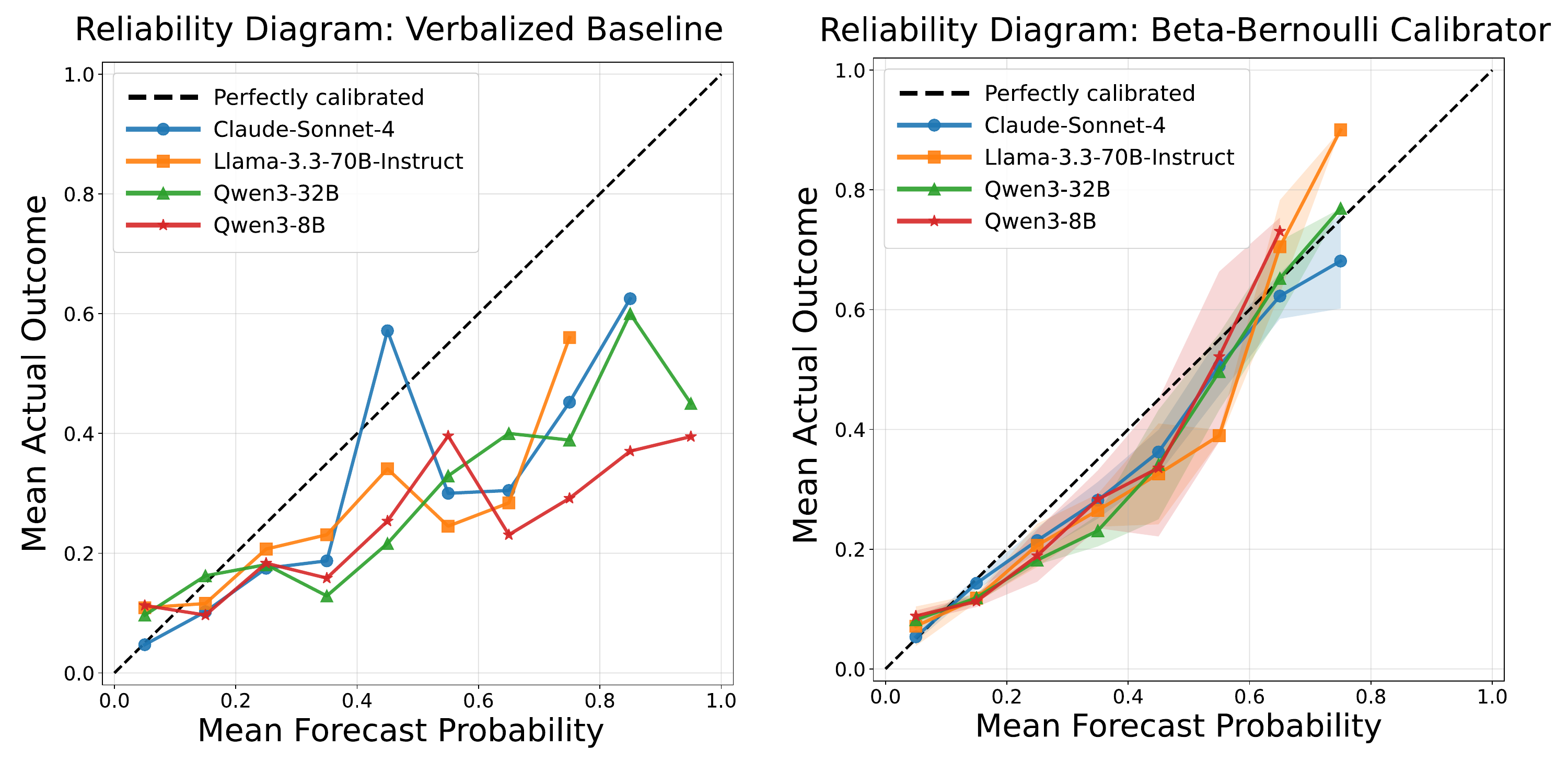}}
    \caption{
      Reliability diagrams. Left: Verbalized probability forecasts are overconfident. Right: Our Beta-Bernoulli Calibrator improves calibration. Bands show $\pm 1$ std across the top-5 runs.
    }
    \label{fig:reliability_diagram_1}
  \end{center}
\end{figure}

As shown in Table \ref{tab:results} and \ref{tab:app_results}, our Beta-Bernoulli Calibrator (BBC) significantly improves over the initial verbalized baseline. For example, using Claude-Sonnet-4 \citep{anthropic2025claude4} as the input LLM, BBC (binary+human) reduces the Brier score from $0.146$ to $0.125$ ($14.4\%$ improvement) and increases AUC from $72.3\%$ to $74.2\%$. Figures \ref{fig:reliability_diagram_1} and \ref{fig:reliability_diagram_2} visualize the calibration gains: while verbalized probability forecasts exhibit overconfidence noted in prior work \citep{schoenegger2024wisdom, halawi2024approaching, nel2025large}, our framework shifts the curves toward the identity line. In general, stronger input LLMs provide better raw forecasts, and BBC effectively builds on these stronger priors without requiring any fine-tuning of the input model. The logit-based $P(\mathrm{True})$ method is poorly calibrated with high ECE. As the LLM generates a rationale before choosing ``Yes'' or ``No'', this intermediate reasoning often amplifies the model’s preference for one outcome, pushing the resulting token probabilities toward extreme values near 0 or 1. Ensembling provides modest improvements over the verbalized baseline, but remains less calibrated than BBC.

Compared to post-hoc calibration baselines (Platt Scaling and Isotonic Regression), BBC consistently achieves better Brier score and stronger discrimination (AUC). While these methods reduce ECE by learning global mappings, they are fundamentally limited by their monotonic nature, which prevents them from improving ranking performance. Notably, our lightweight calibrator exceeds models specifically fine-tuned for forecasting (OpenForecaster-8B and Future-as-a-label-32B). This suggests applying a lightweight calibrator to the base LLM can be a more efficient alternative to fine-tuning the underlying model itself. Moreover, as our framework is model-agnostic, it can be applied on top of any forecasting model, including those fine-tuned forecasters, to further improve performance. Table \ref{tab:finetuned_plus_bbc} shows that BBC may still provide gains beyond traditional calibration methods on most metrics.

\begin{table*}[h]
  \begin{minipage}[t]{0.49\textwidth}
    \caption{Applying BBC (binary+human) on top of forecasting-specialized models further improves forecasts, with consistent gains in Brier score and AUC over other post-hoc calibration methods.}
    \label{tab:finetuned_plus_bbc}
    \centering
    \footnotesize
    \setlength{\tabcolsep}{4pt}
    \renewcommand{\arraystretch}{1}
    \begin{tabular}{@{}l r r r r@{}}
      \toprule
      \tableheader Input LLM / Method &
      \tableheader Brier$\downarrow$ & \tableheader Acc$\uparrow$ &
      \tableheader AUC$\uparrow$ & \tableheader ECE$\downarrow$ \\
      \midrule
      \textbf{OpenForecaster-8B} \\
      \rowwhite \hspace*{1.2em}Verbalized          & 0.157 & 0.794 & 0.663 & 0.084 \\
      \rowwhite \hspace*{1.2em}Platt Scaling       & 0.141 & \best{0.824} & 0.663 & 0.059 \\
      \rowwhite \hspace*{1.2em}Isotonic Regression & \secondbest{0.139} & 0.820 & \secondbest{0.665} & \best{0.044} \\
      \rowblue  \hspace*{1.2em}BBC                 & \best{0.136} & \secondbest{0.821} & \best{0.690} & \secondbest{0.051} \\
      \midrule
      \textbf{Future-as-a-label-32B} \\
      \rowwhite \hspace*{1.2em}Verbalized          & 0.137 & \secondbest{0.829} & \secondbest{0.677} & 0.046 \\
      \rowwhite \hspace*{1.2em}Platt Scaling       & 0.138 & 0.825 & 0.676 & 0.061 \\
      \rowwhite \hspace*{1.2em}Isotonic Regression & \secondbest{0.134} & 0.828 & \secondbest{0.677} & \best{0.037} \\
      \rowblue  \hspace*{1.2em}BBC                 & \best{0.132} & \best{0.833} & \best{0.694} & \secondbest{0.041} \\
      \bottomrule
    \end{tabular}
  \end{minipage}
  \hfill
  \begin{minipage}[t]{0.49\textwidth}
    \caption{OOD performance on the Kalshi dataset. BBC generalizes better than traditional post-hoc calibration methods, and achieves better calibration than forecasting-specialized models.}
    \label{tab:ood_calibration}
    \centering
    \footnotesize
    \setlength{\tabcolsep}{4pt}
    \renewcommand{\arraystretch}{1}
    \begin{tabular}{@{}l r r r r@{}}
      \toprule
      \tableheader Input LLM / Method &
      \tableheader Brier$\downarrow$ & \tableheader Acc$\uparrow$ &
      \tableheader AUC$\uparrow$ & \tableheader ECE$\downarrow$ \\
      \midrule
      \textbf{Qwen3-32B} \\
      \rowwhite \hspace*{1.2em}Verbalized            & \secondbest{0.238} & 0.605 & \secondbest{0.651} & 0.097 \\
      \rowwhite \hspace*{1.2em}Platt Scaling         & 0.251 & 0.596 & \secondbest{0.651} & 0.148 \\
      \rowwhite \hspace*{1.2em}Isotonic Regression   & 0.244 & 0.605 & \secondbest{0.651} & 0.141 \\
      \rowwhite \hspace*{1.2em}Future-as-a-label-32B   & 0.258 & \secondbest{0.607} & 0.638 & 0.159 \\
      \rowblue  \hspace*{1.2em}BBC                   & \best{0.228} & \best{0.609} & \best{0.658} & \best{0.059} \\
      \midrule
      \textbf{Qwen3-8B} \\
      \rowwhite \hspace*{1.2em}Verbalized            & 0.258 & 0.585 & 0.609 & 0.116 \\
      \rowwhite \hspace*{1.2em}Platt Scaling         & 0.258 & 0.595 & 0.609 & 0.146 \\
      \rowwhite \hspace*{1.2em}Isotonic Regression   & 0.258 & 0.597 & 0.608 & 0.152 \\
      \rowwhite \hspace*{1.2em}OpenForecaster-8B     & \secondbest{0.244} & \best{0.599} & \best{0.632} & \secondbest{0.093} \\
      \rowblue  \hspace*{1.2em}BBC                   & \best{0.235} & \best{0.599} & \secondbest{0.620} & \best{0.061} \\
      \bottomrule
    \end{tabular}
  \end{minipage}
\end{table*}

The human baseline remains substantially stronger than current LLM-based forecasters, with Brier score of 0.061 and AUC of 0.958, motivating their use as supervision.\footnote{Computed using the mean crowd forecast.} Comparing BBC (binary only) to BBC (binary+human), we observe further improvements especially in Brier score and AUC. This suggests that human forecast distributions provide a consensus signal about the latent event probability, offering informative supervision beyond a single realized outcome.\footnote{While our training set contains high-quality human forecasts (Brier $=0.085$), for those interested we provide stress test results in Appendix~\ref{app:robustness_human} checking BBC's performance under sparse or biased human supervision.} To further quantify how effectively our framework moves the mixture of Beta distribution closer to the human forecast distribution, we compute the KL divergence between these two. As shown in Table \ref{tab:results}, across all input LLMs, adding human distributional supervision consistently reduces KL divergence, indicating that the learned distributions better match human beliefs.

\section{Analysis and ablations}
\label{sec:analysis_and_ablations}
\subsection{Analysis on epistemic uncertainty}
\label{sec:analysis_u}
We study the epistemic uncertainty in this section. Ideally, when a model is highly uncertain about its own forecast, on average we expect higher prediction error (Brier score). We check this by plotting the Brier score as a function of ranked uncertainty. As discussed in Section \ref{sec:bbc_overview}, we quantify BBC's epistemic uncertainty using the variance of the predicted Beta distribution. We compare against two baselines: (i) verbalized confidence, obtained by prompting the input LLM to report a confidence score after giving an answer, with uncertainty defined as $u=1-\text{confidence}$, and (ii) sampling-based uncertainty, which we take the variance of multiple samples (taken from the ensemble baseline). 

\begin{figure}[h]
  \begin{center}
    \centerline{\includegraphics[width=0.85\columnwidth]{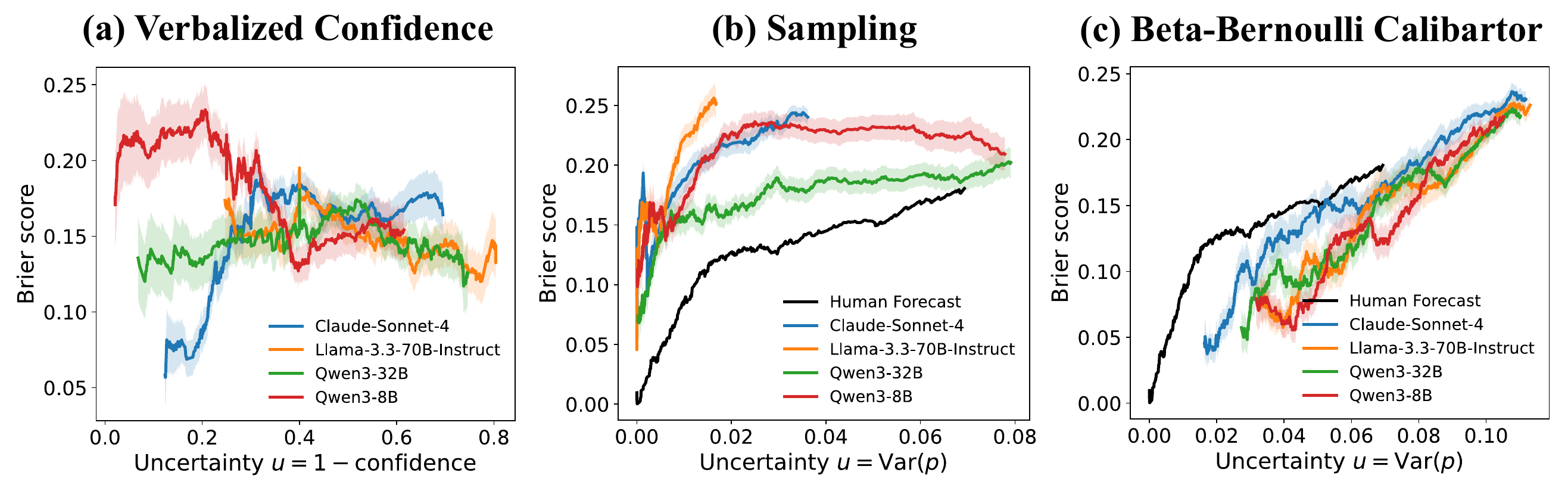}}
    \caption{Brier score vs. ranked epistemic uncertainty, smoothed with a window of 300. (a) Verbalized confidence, (b) Sampling-based variance, and (c) Predicted Beta distribution variance.}
    \label{fig:u_vs_e_1}
    
  \end{center}
\vskip -0.2in
\end{figure}

Figure \ref{fig:u_vs_e_1}(a) shows that the self-reported confidence is noisy and disjointed from empirical performance: lower verbalized confidence in general does not correspond to a lower Brier score. The sampling-based uncertainty in Figure \ref{fig:u_vs_e_1}(b) is more informative, but it becomes less discriminative at higher uncertainty. In contrast, in Figure \ref{fig:u_vs_e_1}(c), BBC produces an uncertainty measure that is consistently aligned with errors across input LLMs, similar to the trend in the human forecast baseline, offering a more reliable signal of forecasting errors.

% \vskip -0.1in
\subsection{Generalization to out-of-distribution data}
To further test if our calibrator is robust in the out-of-distribution (OOD) setting, we evaluate it on questions from the prediction platform Kalshi. After applying similar topic and volume filtering as in our main dataset, we collect 3,208 questions that resolved after August 2025.\footnote{See details in Appendix \ref{app:dataset}.} Table \ref{tab:ood_calibration} shows that traditional post-hoc calibration methods fail to generalize well when tested on this dataset, resulting in even worse calibration with higher ECE. In contrast, BBC maintains strong performance, achieving better calibration and discrimination performance. It even outperforms the forecasting-specialized models in Brier score and ECE.

\begin{table*}[h]
  \begin{minipage}[t]{0.49\textwidth}
    \caption{Ablation on calibrator family and size for BBC. 1B calibrator is already an effective choice.
    }
    \label{tab:ablation_model_size}
    \centering
    \footnotesize
    \setlength{\tabcolsep}{4pt}
    \renewcommand{\arraystretch}{1}
    \begin{tabular}{@{}l r r r r@{}}
      \toprule
      \tableheader Input LLM / Calibrator &
      \tableheader Brier$\downarrow$ & \tableheader Acc$\uparrow$ &
      \tableheader AUC$\uparrow$ & \tableheader ECE$\downarrow$ \\
      \midrule
      \textbf{Claude-Sonnet-4} \\
      \rowwhite  \hspace*{1.2em}Llama-3.2-1B       & \secondbest{0.126} & \secondbest{0.836} & \secondbest{0.744} & \secondbest{0.036} \\
      \rowwhite \hspace*{1.2em}Qwen2.5-0.5B       & 0.129 & 0.827 & 0.737 & 0.044 \\
      \rowwhite \hspace*{1.2em}Llama-3.2-3B       & \best{0.124} & \best{0.840} & 0.741 & \best{0.030} \\
      \rowwhite \hspace*{1.2em}Qwen3-4B-Instruct  & 0.127 & 0.835 & 0.736 & 0.037 \\
      \rowwhite \hspace*{1.2em}Llama-3.1-8B       & \best{0.124} & 0.834 & \best{0.752} & \secondbest{0.036} \\
      \midrule
      \textbf{Llama-3.3-70B-Instruct} \\
      \rowwhite  \hspace*{1.2em}Llama-3.2-1B       & 0.136 & \best{0.830} & \secondbest{0.679} & 0.060 \\
      \rowwhite \hspace*{1.2em}Qwen2.5-0.5B       & 0.138 & 0.818 & 0.667 & 0.061 \\
      \rowwhite \hspace*{1.2em}Llama-3.2-3B       & \secondbest{0.135} & \best{0.830} & 0.677 & 0.059 \\
      \rowwhite \hspace*{1.2em}Qwen3-4B-Instruct  & 0.138 & 0.821 & 0.675 & \best{0.045} \\
      \rowwhite \hspace*{1.2em}Llama-3.1-8B       & \best{0.134} & \secondbest{0.826} & \best{0.693} & \secondbest{0.058} \\
      \bottomrule
    \end{tabular}
  \end{minipage}
  \hfill
  \begin{minipage}[t]{0.49\textwidth}
    \caption{Ablation on calibrator input. Removing the initial forecast drops performance, while adding a rationale brings minimal benefit at extra cost.}
    \label{tab:ablation_input}
    \centering
    \footnotesize
    \setlength{\tabcolsep}{4pt}
    \renewcommand{\arraystretch}{1}
    \begin{tabular}{@{}l r r r r@{}}
      \toprule
      \tableheader Setting / Method &
      \tableheader Brier$\downarrow$ & \tableheader Acc$\uparrow$ &
      \tableheader AUC$\uparrow$ & \tableheader ECE$\downarrow$ \\
      \midrule
      \rowwhite \textbf{BBC w/o initial forecast} & 0.140 & 0.827 & 0.650 & 0.060 \\
      \midrule
      \textbf{Claude-Sonnet-4} \\
      \rowwhite  \hspace*{1.2em}BBC w initial forecast & 0.126 & 0.836 & 0.744 & \best{0.036} \\
      \rowwhite \hspace*{1.2em}+ Reasoning            & \best{0.125} & \best{0.837} & \best{0.745} & 0.041 \\
      \midrule
      \textbf{Llama-3.3-70B-Instruct} \\
      \rowwhite  \hspace*{1.2em}BBC w initial forecast & \best{0.136} & \best{0.830} & 0.679 & 0.060 \\
      \rowwhite \hspace*{1.2em}+ Reasoning            & 0.137 & 0.821 & \best{0.682} & \best{0.052} \\
      \bottomrule
    \end{tabular}
  \end{minipage}
  \vskip -0.15in
\end{table*}

\subsection{Ablation: calibrator model family and size}
\label{sec:ablation_model_size}
In our main experiments, we use Llama-3.2-1B as the base model for BBC, and show that it is already an efficient choice that outperforms standard baselines. Here we analyze the effect of varying both the calibrator family and size, as shown in Table \ref{tab:ablation_model_size}. Overall, scaling the calibrator provides only modest changes in Brier and accuracy, but we observe a clear performance gain in AUC with a larger 8B model, showing the potential in scaling up. Comparing model families, Llama-based calibrators consistently outperform similarly sized Qwen-based models.

\subsection{Ablation: input content}
Table \ref{tab:ablation_input} studies how the calibrator input $x_i$ affects BBC. For our main experiments, $x_i$ encodes both the event information and an initial forecast $\hat p ^ \text{init}$. When removing $\hat p ^ \text{init}$,  BBC no longer acts as a post-hoc calibrator and a significant drop in AUC is observed. This indicates that BBC is most effective when refining an existing belief, benefiting from the stronger input LLM forecast rather than predicting from scratch. However, further enriching $x_i$ yields only marginal gains while introducing additional computational overhead. In particular, appending the input LLM’s rationale results in only a slight increase in AUC at higher computational cost, suggesting that providing the initial forecast alone is sufficient in practice.

\section{Conclusion}
We introduce the Beta-Bernoulli Calibrator, a lightweight and model-agnostic post-hoc calibration method that learns from both binary outcomes and the distribution of human forecasts. Our model maps from an initial probability forecast to a Beta distribution over event likelihood. Across multiple input LLMs, we show that the Beta mean provides a better calibrated and more accurate point forecast, and the Beta variance serves as a measure of epistemic uncertainty that is predictive of forecasting errors. Moreover, BBC demonstrates consistently better calibration than models specifically fine-tuned for forecasting, observed both in- and out-of-distribution.

\section*{Acknowledgments}
This work was supported in part by Visko AI, Toyota Research Institute R2I program, a Google TPU Award, and the Institute of Information \& Communications Technology Planning Evaluation (IITP) under grant RS-2024-00469482, funded by the Ministry of Science and ICT (MSIT) of the Republic of Korea in connection with the Global AI Frontier Lab International Collaborative Research. The compute is supported by the NYU High Performance Computing resources, services, and staff expertise.

\bibliographystyle{apalike}
\bibliography{ref}

@inproceedings{guo2017calibration,
  title={On calibration of modern neural networks},
  author={Guo, Chuan and Pleiss, Geoff and Sun, Yu and Weinberger, Kilian Q},
  booktitle={ICML},
  year={2017},
}

@article{platt1999probabilistic,
  title={Probabilistic outputs for support vector machines and comparisons to regularized likelihood methods},
  author={Platt, John and others},
  journal={Advances in large margin classifiers},
  volume={10},
  number={3},
  pages={61--74},
  year={1999},
}

@inproceedings{zadrozny2002transforming,
  title={Transforming classifier scores into accurate multiclass probability estimates},
  author={Zadrozny, Bianca and Elkan, Charles},
  booktitle={ACM SIGKDD},
  year={2002}
}

@article{degroot1983comparison,
  title={The comparison and evaluation of forecasters},
  author={DeGroot, Morris H and Fienberg, Stephen E},
  journal={Journal of the Royal Statistical Society: Series D (The Statistician)},
  volume={32},
  number={1-2},
  pages={12--22},
  year={1983},
  publisher={Wiley Online Library}
}

@inproceedings{niculescu2005predicting,
  title={Predicting good probabilities with supervised learning},
  author={Niculescu-Mizil, Alexandru and Caruana, Rich},
  booktitle={ICML},
  year={2005}
}

@inproceedings{zadrozny2001obtaining,
  title={Obtaining calibrated probability estimates from decision trees and naive bayesian classifiers},
  author={Zadrozny, Bianca and Elkan, Charles},
  booktitle={ICML},
  year={2001}
}

@article{shorinwa2025survey,
  title={A survey on uncertainty quantification of large language models: Taxonomy, open research challenges, and future directions},
  author={Shorinwa, Ola and Mei, Zhiting and Lidard, Justin and Ren, Allen Z and Majumdar, Anirudha},
  journal={ACM Computing Surveys},
  volume={58},
  number={3},
  pages={1--38},
  year={2025},
  publisher={ACM New York, NY}
}

@inproceedings{xiong2024can,
  title={Can {LLM}s Express Their Uncertainty? An Empirical Evaluation of Confidence Elicitation in {LLM}s},
  author={Xiong, Miao and Hu, Zhiyuan and Lu, Xinyang and Li, Yifei and Fu, Jie and He, Junxian and Hooi, Bryan},
  booktitle={ICLR},
  year={2024}
}

@inproceedings{tian2023just,
  title={Just Ask for Calibration: Strategies for Eliciting Calibrated Confidence Scores from Language Models Fine-Tuned with Human Feedback},
  author={Tian, Katherine and Mitchell, Eric and Zhou, Allan and Sharma, Archit and Rafailov, Rafael and Yao, Huaxiu and Finn, Chelsea and Manning, Christopher D},
  booktitle={EMNLP},
  year={2023}
}

@inproceedings{mei2026reasoning,
  title={Reasoning about Uncertainty: Do Reasoning Models Know When They Don’t Know?},
  author={Mei, Zhiting and Zhang, Christina and Yin, Tenny and Lidard, Justin and Sho, Ola and Majumdar, Anirudha},
  booktitle={Findings of EACL},
  year={2026}
}

@article{kirichenko2025abstentionbench,
  title={{AbstentionBench}: Reasoning {LLM}s Fail on Unanswerable Questions},
  author={Kirichenko, Polina and Ibrahim, Mark and Chaudhuri, Kamalika and Bell, Samuel J},
  journal={arXiv preprint arXiv:2506.09038},
  year={2025}
}

@article{kadavath2022language,
  title={Language models (mostly) know what they know},
  author={Kadavath, Saurav and Conerly, Tom and Askell, Amanda and Henighan, Tom and Drain, Dawn and Perez, Ethan and Schiefer, Nicholas and Hatfield-Dodds, Zac and DasSarma, Nova and Tran-Johnson, Eli and others},
  journal={arXiv preprint arXiv:2207.05221},
  year={2022}
}

@inproceedings{kapoor2024large,
  title={Large language models must be taught to know what they don’t know},
  author={Kapoor, Sanyam and Gruver, Nate and Roberts, Manley and Collins, Katie and Pal, Arka and Bhatt, Umang and Weller, Adrian and Dooley, Samuel and Goldblum, Micah and Wilson, Andrew G},
  booktitle={NeurIPS},
  year={2024}
}

@inproceedings{ling2024uncertainty,
  title={Uncertainty Quantification for In-Context Learning of Large Language Models},
  author={Ling, Chen and Zhao, Xujiang and Zhang, Xuchao and Cheng, Wei and Liu, Yanchi and Sun, Yiyou and Oishi, Mika and Osaki, Takao and Matsuda, Katsushi and Ji, Jie and others},
  booktitle={NAACL},
  year={2024}
}

@inproceedings{fadeeva2024fact,
  title={Fact-Checking the Output of Large Language Models via Token-Level Uncertainty Quantification},
  author={Fadeeva, Ekaterina and Rubashevskii, Aleksandr and Shelmanov, Artem and Petrakov, Sergey and Li, Haonan and Mubarak, Hamdy and Tsymbalov, Evgenii and Kuzmin, Gleb and Panchenko, Alexander and Baldwin, Timothy and others},
  booktitle={Findings of ACL},
  year={2024}
}

@inproceedings{orgad2025llms,
  title={{LLM}s Know More Than They Show: On the Intrinsic Representation of {LLM} Hallucinations},
  author={Orgad, Hadas and Toker, Michael and Gekhman, Zorik and Reichart, Roi and Szpektor, Idan and Kotek, Hadas and Belinkov, Yonatan},
  booktitle={ICLR},
  year={2025}
}

@inproceedings{zhang2025reasoning,
  title={Reasoning Models Know When They're Right: Probing Hidden States for Self-Verification},
  author={Zhang, Anqi and Chen, Yulin and Pan, Jane and Zhao, Chen and Panda, Aurojit and Li, Jinyang and He, He},
  booktitle={COLM},
  year={2025}
}

@inproceedings{azaria2023internal,
  title={The Internal State of an {LLM} Knows When It’s Lying},
  author={Azaria, Amos and Mitchell, Tom},
  booktitle={Findings of EMNLP},
  year={2023}
}

@article{lin2022teaching,
  title={Teaching Models to Express Their Uncertainty in Words},
  author={Lin, Stephanie and Hilton, Jacob and Evans, Owain},
  journal={Transactions on Machine Learning Research},
  year={2022}
}

@inproceedings{li2025conftuner,
  title={{ConfTuner}: Training Large Language Models to Express Their Confidence Verbally},
  author={Li, Yibo and Xiong, Miao and Wu, Jiaying and Hooi, Bryan},
  booktitle={NeurIPS},
  year={2025}
}

@inproceedings{damani2026beyond,
  title={Beyond binary rewards: Training {LM}s to reason about their uncertainty},
  author={Damani, Mehul and Puri, Isha and Slocum, Stewart and Shenfeld, Idan and Choshen, Leshem and Kim, Yoon and Andreas, Jacob},
  booktitle={ICLR},
  year={2026}
}

@inproceedings{xu2024sayself,
  title={{SaySelf}: Teaching {LLM}s to Express Confidence with Self-Reflective Rationales},
  author={Xu, Tianyang and Wu, Shujin and Diao, Shizhe and Liu, Xiaoze and Wang, Xingyao and Chen, Yangyi and Gao, Jing},
  booktitle={EMNLP},
  year={2024}
}

@inproceedings{jiang2023calibrating,
  title={Calibrating language models via augmented prompt ensembles},
  author={Jiang, Mingjian and Ruan, Yangjun and Huang, Sicong and Liao, Saifei and Pitis, Silviu and Grosse, Roger Baker and Ba, Jimmy},
  booktitle={ICML},
  year={2023}
}

@inproceedings{zhang2024luq,
  title={{LUQ}: Long-text Uncertainty Quantification for {LLM}s},
  author={Zhang, Caiqi and Liu, Fangyu and Basaldella, Marco and Collier, Nigel},
  booktitle={EMNLP},
  year={2024}
}

@inproceedings{karger2024forecastbench,
  title={{ForecastBench}: A Dynamic Benchmark of {AI} Forecasting Capabilities},
  author={Karger, Ezra and Bastani, Houtan and Yueh-Han, Chen and Jacobs, Zachary and Halawi, Danny and Zhang, Fred and Tetlock, Philip},
  booktitle={ICLR},
  year={2024}
}

@article{zeng2025futurex,
  title={{FutureX}: An advanced live benchmark for {LLM} agents in future prediction},
  author={Zeng, Zhiyuan and Liu, Jiashuo and Chen, Siyuan and He, Tianci and Liao, Yali and Tian, Yixiao and Wang, Jinpeng and Wang, Zaiyuan and Yang, Yang and Yin, Lingyue and others},
  journal={arXiv preprint arXiv:2508.11987},
  year={2025}
}

@inproceedings{yang2026llm,
  title={{LLM}-as-a-Prophet: Understanding Predictive Intelligence with {Prophet Arena}},
  author={Yang, Qingchuan and Mahns, Simon and Li, Sida and Gu, Anri and Wu, Jibang and Xu, Haifeng},
  booktitle={ICLR},
  year={2026}
}

@inproceedings{halawi2024approaching,
  title={Approaching human-level forecasting with language models},
  author={Halawi, Danny and Zhang, Fred and Yueh-Han, Chen and Steinhardt, Jacob},
  booktitle={NeurIPS},
  year={2024}
}

@article{chandak2025scaling,
  title={Scaling Open-Ended Reasoning to Predict the Future},
  author={Chandak, Nikhil and Goel, Shashwat and Prabhu, Ameya and Hardt, Moritz and Geiping, Jonas},
  journal={arXiv preprint arXiv:2512.25070},
  year={2025}
}

@article{nel2025large,
  title={Do Large Language Models Know What They Don't Know? {Kalshibench}: A New Benchmark for Evaluating Epistemic Calibration via Prediction Markets},
  author={Nel, Lukas},
  journal={arXiv preprint arXiv:2512.16030},
  year={2025}
}

@article{alur2025aia,
  title={{AIA} Forecaster: Technical Report},
  author={Alur, Rohan and Stadie, Bradly C and Kang, Daniel and Chen, Ryan and McManus, Matt and Rickert, Michael and Lee, Tyler and Federici, Michael and Zhu, Richard and Fogerty, Dennis and others},
  journal={arXiv preprint arXiv:2511.07678},
  year={2025}
}

@article{schoenegger2024wisdom,
  title={Wisdom of the silicon crowd: {LLM} ensemble prediction capabilities rival human crowd accuracy},
  author={Schoenegger, Philipp and Tuminauskaite, Indre and Park, Peter S and Bastos, Rafael Valdece Sousa and Tetlock, Philip E},
  journal={Science Advances},
  volume={10},
  number={45},
  pages={eadp1528},
  year={2024},
  publisher={American Association for the Advancement of Science}
}

@article{cobbe2021training,
  title={Training verifiers to solve math word problems},
  author={Cobbe, Karl and Kosaraju, Vineet and Bavarian, Mohammad and Chen, Mark and Jun, Heewoo and Kaiser, Lukasz and Plappert, Matthias and Tworek, Jerry and Hilton, Jacob and Nakano, Reiichiro and others},
  journal={arXiv preprint arXiv:2110.14168},
  year={2021}
}

@inproceedings{yang2018hotpotqa,
  title={{HotpotQA}: A dataset for diverse, explainable multi-hop question answering},
  author={Yang, Zhilin and Qi, Peng and Zhang, Saizheng and Bengio, Yoshua and Cohen, William and Salakhutdinov, Ruslan and Manning, Christopher D},
  booktitle={EMNLP},
  year={2018}
}

@inproceedings{kendall2017uncertainties,
  title={What uncertainties do we need in bayesian deep learning for computer vision?},
  author={Kendall, Alex and Gal, Yarin},
  booktitle={NeurIPS},
  year={2017}
}

@book{tetlock2016superforecasting,
  title={Superforecasting: The art and science of prediction},
  author={Tetlock, Philip E and Gardner, Dan},
  year={2016},
  publisher={Random House}
}

@incollection{lahiri2013forecasting,
  title={Forecasting binary outcomes},
  author={Lahiri, Kajal and Yang, Liu},
  booktitle={Handbook of economic forecasting},
  volume={2},
  pages={1025--1106},
  year={2013},
  publisher={Elsevier}
}

@article{brier1950verification,
  title={Verification of forecasts expressed in terms of probability},
  author={Brier, Glenn W},
  journal={Monthly weather review},
  volume={78},
  number={1},
  pages={1--3},
  year={1950},
  publisher={War Department, Office of the Chief Signal Officer}
}

@article{bradley1997use,
  title={The use of the area under the ROC curve in the evaluation of machine learning algorithms},
  author={Bradley, Andrew P},
  journal={Pattern recognition},
  volume={30},
  number={7},
  pages={1145--1159},
  year={1997},
  publisher={Elsevier}
}

@article{turtel2026future,
  title={{Future-as-Label}: Scalable Supervision from Real-World Outcomes},
  author={Turtel, Benjamin and Wilczewski, Paul and Franklin, Danny and Skothiem, Kris},
  journal={arXiv preprint arXiv:2601.06336},
  year={2026}
}

@inproceedings{naeini2015obtaining,
  title={Obtaining well calibrated probabilities using bayesian binning},
  author={Naeini, Mahdi Pakdaman and Cooper, Gregory and Hauskrecht, Milos},
  booktitle={AAAI},
  year={2015}
}

@article{gneiting2007strictly,
  title={Strictly proper scoring rules, prediction, and estimation},
  author={Gneiting, Tilmann and Raftery, Adrian E},
  journal={Journal of the American statistical Association},
  volume={102},
  number={477},
  pages={359--378},
  year={2007},
  publisher={Taylor \& Francis}
}

@article{grattafiori2024llama,
  title={The llama 3 herd of models},
  author={Grattafiori, Aaron and Dubey, Abhimanyu and Jauhri, Abhinav and Pandey, Abhinav and Kadian, Abhishek and Al-Dahle, Ahmad and Letman, Aiesha and Mathur, Akhil and Schelten, Alan and Vaughan, Alex and others},
  journal={arXiv preprint arXiv:2407.21783},
  year={2024}
}

@inproceedings{hu2022lora,
  title={{LoRA}: Low-rank adaptation of large language models},
  author={Hu, Edward J and Shen, Yelong and Wallis, Phillip and Allen-Zhu, Zeyuan and Li, Yuanzhi and Wang, Shean and Wang, Lu and Chen, Weizhu and others},
  booktitle={ICLR},
  year={2022}
}

@misc{anthropic2025claude4,
  title        = {Introducing Claude 4},
  author       = {{Anthropic}},
  year         = {2025},
  howpublished = {\url{https://www.anthropic.com/news/claude-4}},
}

@article{yang2025qwen3,
  title={Qwen3 technical report},
  author={Yang, An and Li, Anfeng and Yang, Baosong and Zhang, Beichen and Hui, Binyuan and Zheng, Bo and Yu, Bowen and Gao, Chang and Huang, Chengen and Lv, Chenxu and others},
  journal={arXiv preprint arXiv:2505.09388},
  year={2025}
}

@article{hurst2024gpt,
  title={{GPT}-4o system card},
  author={Hurst, Aaron and Lerer, Adam and Goucher, Adam P and Perelman, Adam and Ramesh, Aditya and Clark, Aidan and Ostrow, AJ and Welihinda, Akila and Hayes, Alan and Radford, Alec and others},
  journal={arXiv preprint arXiv:2410.21276},
  year={2024}
}

@article{qwen2025qwen25technicalreport,
      title={Qwen2.5 Technical Report}, 
      author={Qwen Team},
      journal={arXiv preprint arXiv:2412.15115},
      year={2025}
}

@inproceedings{sensoy2018evidential,
  title={Evidential deep learning to quantify classification uncertainty},
  author={Sensoy, Murat and Kaplan, Lance and Kandemir, Melih},
  booktitle={NeurIPS},
  year={2018}
}

@inproceedings{charpentier2020posterior,
  title={Posterior network: Uncertainty estimation without ood samples via density-based pseudo-counts},
  author={Charpentier, Bertrand and Z{\"u}gner, Daniel and G{\"u}nnemann, Stephan},
  booktitle={NeurIPS},
  year={2020}
}

@article{murphy2026agentic,
  title={Agentic Forecasting using Sequential Bayesian Updating of Linguistic Beliefs},
  author={Murphy, Kevin},
  journal={arXiv preprint arXiv:2604.18576},
  year={2026}
}

\newpage
\appendix
\section*{Appendix}

\section{Limitations}
\label{app:limitations}
Our work has several limitations. First, as we gather human forecasts from prediction platforms, the training signal may inherit topic biases to politics and economics. Second, although our ablations indicate that larger calibrators are possible to bring further gains, we do not fully explore this scaling trend and leave a systematic study of the calibrator's upper bound to future work. Third, BBC does not incorporate information updates, and instead relies solely on its internal knowledge to calibrate the belief. A natural direction for future work is to extend BBC to condition on the temporal dimension and intermediate evidence, enabling us to model how human uncertainty shifts over time when new information emerges.

\section{Evaluation metrics}
\label{app:metrics}

\paragraph{Brier score} \citep{brier1950verification} is the Mean Squared Error between the predicted probabilities and the binary outcomes. A lower Brier score corresponds to better predictions.

$$\text{BS} = \frac{1}{N} \sum_{i=1}^N (\hat{p}_i - y_i)^2.$$

\paragraph{Accuracy} is the fraction of events predicted correctly, given that an event is predicted positive when the predicted probability exceeds a threshold (e.g. $0.5$).

$$
\text{Acc} = \frac{1}{N}\sum_{i=1}^N \mathbb{I}\{(\hat{p_i} \ge 0.5) = y_i\}.
$$

\paragraph{AUC} \citep{bradley1997use} is defined as the Area Under the Receiver Operating Characteristic curve. It is independent of any threshold, measuring how well $\hat{p}_i$ can discriminate events that happen versus those do not.

\paragraph{Expected Calibration Error (ECE)} \citep{naeini2015obtaining} is a calibration metric. Ideally, for example, if we track 100 events where the model predicts a probability of $0.7$, we expect to see 70 events occur in the end. The perfect calibration is formally written by $P(y=1 \mid p = q) = q, \forall q$. ECE measures how uncalibrated a model is by taking the expectation of the absolute difference between the model prediction and the empirical event occurrences, computed from the set of events with similar predictions. More precisely, we split the events into equal bins based on the model predictions (e.g., $[0, 0.1], (0.1, 0.2], \dots$). For each bin, we compute (i) the average predicted probability $\text{prob}(B_m)$, and (ii) the empirical accuracy $\text{acc}(B_m)$ -- the fraction of events in this bin being true. Then, ECE is defined as the weighted average of absolute differences in these two quantities:

$$\text{ECE} = \sum_{m=1}^M \frac{|B_m|}{N} \left| \text{acc}(B_m) - \text{prob}(B_m) \right|$$
$$
    \text{where prob}(B_m) = \frac{\sum_{i \in B_m} \hat{p}_i}{|B_m|} , \text{ acc}(B_m) = \frac{\sum_{i \in B_m} y_i}{|B_m|} .
$$

Moreover, visualizing $\text{acc}(B_m)$ against $\text{prob}(B_m)$ produces a \textbf{Reliability Diagram} \citep{degroot1983comparison, niculescu2005predicting}, where a perfectly calibrated model follows the identity line $y=x$.

\section{Why human forecasts help: a toy experiment}
\label{app:toy}
 
In this section, we provide a toy experiment to further demonstrate the necessity of human forecasts as distributional supervision. The toy experiment aims to show that, if the aggregated human forecast distribution is an approximate distribution over the latent event probability, it can provide both the missing signal about the distribution shape, and additional supervision beyond the single binary outcome, resulting in better forecasts.
 
We construct a synthetic Beta-Bernoulli setting. We generate $30{,}000$ questions with $10$-dimensional input features mapped through a nonlinear function into three distinct ground-truth regimes: (i) \emph{Confident YES:} $\text{Beta}(50,10)$ with $p=0.83$; (ii) \emph{Uncertain:} $\text{Beta}(5,5)$ with $p=0.5$; and (iii)  \emph{Confident NO:} $\text{Beta}(10,50)$ with $p=0.17$. Each question receives a single binary outcome $y \sim \text{Bernoulli}(p)$ with $p \sim \text{Beta}(\alpha, \beta)$. Human forecasts are simulated by drawing $1,000$ samples from the true Beta distribution. We train a 2-layer MLP that maps input features to Beta parameters $(\alpha, \beta)$ under three loss configurations: \emph{Binary only} (BCE), \emph{Human only}, and \emph{Binary + Human}.
 
\begin{figure}[h]
  \begin{center}
\centerline{\includegraphics[width=\columnwidth]{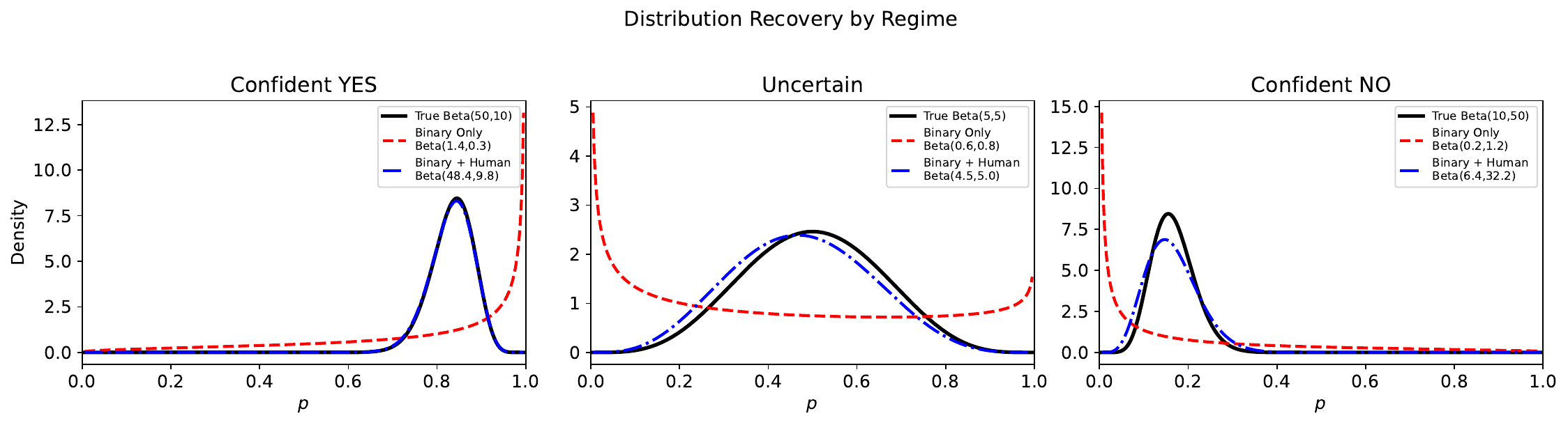}}
    \caption{Toy experiment: predicted Beta distributions vs. ground truth across the three regimes. Binary-only training (red dashed) fails to recover the distribution shape, while binary+human (blue) approximately recovers the ground truth (black).}
    \label{fig:toy_distributions}
  \end{center}
   \vskip -0.3in
\end{figure}

We can see that binary-only training achieves a reasonable Brier score of $0.194$ (Table~\ref{tab:toy_metrics}) and learns approximate means (Table~\ref{tab:toy_recovery}), but completely fails to recover the Beta shape: the predicted distributions (red) deviate substantially from the ground truth (black) in Figure~\ref{fig:toy_distributions}. Adding human supervision recovers parameters close to ground truth (Figure \ref{fig:toy_distributions}) and provides better forecasting performance (Tables~\ref{tab:toy_metrics}).
 
Therefore, while BCE is optimal for estimating the mean event probability given sufficient repeated observations, forecasting settings provide only a single realization per event. Human forecast distributions act as a noisy proxy for the latent probability distribution and provide additional information about uncertainty that improves calibration beyond binary supervision alone. Empirically, this holds as long as human forecasts are a reasonable proxy of the true underlying distribution. In our training set, human forecasts significantly outperform all LLMs (human: Brier $=0.085$, AUC $=0.945$ vs. model average: Brier $=0.196$, AUC $=0.704$), validating their quality as supervision.

\begin{table}[h]
  \caption{Toy experiment: forecasting metrics across three loss configurations. Best results are bolded.}
  \label{tab:toy_metrics}
  \begin{center}
    \small
    \renewcommand{\arraystretch}{1}
    \begin{tabular}{lrrrr}
      \toprule
      \tableheader Method & \tableheader Brier$\downarrow$ & \tableheader Acc$\uparrow$ &
      \tableheader AUC$\uparrow$ & \tableheader ECE$\downarrow$ \\
      \midrule
      \rowwhite Binary only      & 0.194        & 0.709        & 0.775        & 0.053 \\
      \rowwhite Human only       & \best{0.183} & 0.717        & 0.780        & 0.016 \\
      \rowwhite  Binary + Human   & \best{0.183} & \best{0.719} & \best{0.784} & \best{0.011} \\
      \bottomrule
    \end{tabular}
  \end{center}
\end{table}

\begin{table}[h]
  \caption{Toy experiment: recovery of ground-truth Beta parameters. Binary-only training fails to recover $(\alpha, \beta)$ while binary+human recovers parameters close to ground truth.}
  \label{tab:toy_recovery}
  \begin{center}
    \small
    \setlength{\tabcolsep}{5pt}
    \renewcommand{\arraystretch}{1}
    \begin{tabular}{l cc cc cc}
      \toprule
      & \multicolumn{2}{c}{\tableheader Ground truth} & \multicolumn{2}{c}{\tableheader Binary only} & \multicolumn{2}{c}{\tableheader Binary + Human} \\
      \cmidrule(lr){2-3}\cmidrule(lr){4-5}\cmidrule(lr){6-7}
      \tableheader Regime & \tableheader $(\alpha, \beta)$ & \tableheader Mean &
      \tableheader $(\alpha, \beta)$ & \tableheader Mean &
      \tableheader $(\alpha, \beta)$ & \tableheader Mean \\
      \midrule
      \rowwhite Confident YES   & $(50, 10)$ & 0.833 & $(1.4, 0.4)$ & 0.778 & \best{$(44.8,\ 9.1)$}  & 0.831 \\
      \rowwhite Uncertain       & $(5, 5)$   & 0.500 & $(0.6, 0.8)$ & 0.429 & \best{$(4.5,\ 5.0)$}   & 0.474 \\
      \rowwhite Confident NO    & $(10, 50)$ & 0.167 & $(0.2, 1.2)$ & 0.143 & \best{$(6.7,\ 33.5)$}  & 0.167 \\
      \bottomrule
    \end{tabular}
  \end{center}
\end{table}

\section{Dataset}
\label{app:dataset}
\subsection{Data preprocessing}
\paragraph{Metaculus.} We obtain Metaculus data from the public API \url{https://www.metaculus.com/api2/questions/}. We filter to binary questions with at least one human forecast, and exclude meta-questions that predict the community prediction on another Metaculus question.
\paragraph{Polymarket.} For Polymarket, we exclude the sports, cryptocurrency, and weather domains. The question open date is set to be the earlier of (i) 30 days before the last observed timestamp and (ii) 7 days after the first observed timestamp. We notice the number of recently resolved questions is much larger than in earlier periods, which would make the test set disproportionately large. To address this, we filter by popularity rather than random sampling. Specifically, we filter for questions with at least 5 price history entries for training, at least 30 for validation, and at least 100 for testing. 
\paragraph{Kalshi.} For the OOD Kalshi dataset, we again exclude the sports, cryptocurrency, and weather domains. We further filter for events that have a total trading volume greater than 10,000. This results in 3,208 events that resolved after August 2025. Since Kalshi events can include multiple markets (outcome options), we further convert them to binary questions by randomly selecting one market per event and asking whether that outcome occurs. 

\subsection{Dataset statistics}
\label{app:dataset_statistics}

Table \ref{tab:data_stat} provides the dataset statistics by source and split. Table \ref{tab:category_dist_main} presents the category distribution. Most questions in our main dataset fall under ``Politics \& Governance'' (6,163) and ``Economics \& Business'' (1,960), two domains closely tied to real-world decision-making. We use GPT-4o-mini \citep{hurst2024gpt} to assign categories to questions, using the prompt from \citet{halawi2024approaching}. Table \ref{tab:category_dist_kalshi} shows the category distribution of Kalshi dataset, using their existing category tags.

\begin{table}[h]
  \caption{Dataset statistics by source and split. Train: resolved before April 2025; Val: resolved between April and July 2025; Test: resolved between August 2025 and January 2026.}
  \label{tab:data_stat}
  \begin{center}
    \small
    \renewcommand{\arraystretch}{1}
    \begin{tabular}{lrrr}
      \toprule
      \tableheader Source & \tableheader Train & \tableheader Val & \tableheader Test \\
      \midrule
      \rowwhite Metaculus  & 3{,}264 & 545     & 420     \\
      \rowwhite Polymarket & 4{,}560 & 1{,}372 & 1{,}194 \\
      \midrule
      \rowwhite Total      & 7{,}824 & 1{,}917 & 1{,}614 \\
      \bottomrule
    \end{tabular}
  \end{center}
\end{table}

\begin{table}[h]
  \begin{minipage}[t]{0.59\textwidth}
    \caption{Category distribution for our main dataset.}
    \label{tab:category_dist_main}
    \centering
    \small
    \setlength{\tabcolsep}{4pt}
    \renewcommand{\arraystretch}{1}
    \begin{tabular}{lrrr}
      \toprule
      \tableheader Category & \tableheader Metaculus & \tableheader Polymarket & \tableheader Total \\
      \midrule
      \rowwhite Politics \& Governance   & 1{,}693 & 4{,}470 & 6{,}163 \\
      \rowwhite Economics \& Business    &   899   & 1{,}061 & 1{,}960 \\
      \rowwhite Arts \& Recreation       &   116   &   762   &   878   \\
      \rowwhite Security \& Defense      &   391   &   238   &   629   \\
      \rowwhite Science \& Tech          &   304   &   185   &   489   \\
      \rowwhite Sports                   &   213   &   226   &   439   \\
      \rowwhite Healthcare \& Biology    &   296   &    44   &   340   \\
      \rowwhite Environment \& Energy    &   194   &    39   &   233   \\
      \rowwhite Other                    &    75   &    91   &   166   \\
      \rowwhite Education \& Research    &    48   &    10   &    58   \\
      \midrule
      \rowwhite \textbf{Total}           & 4{,}229 & 7{,}126 & 11{,}355 \\
      \bottomrule
    \end{tabular}
  \end{minipage}
  \hfill
  \begin{minipage}[t]{0.4\textwidth}
    \caption{Category distribution for the Kalshi dataset.}
    \label{tab:category_dist_kalshi}
    \centering
    \small
    \setlength{\tabcolsep}{5pt}
    \renewcommand{\arraystretch}{1}
    \begin{tabular}{lr}
      \toprule
      \tableheader Category & \tableheader Kalshi \\
      \midrule
      \rowwhite Financials                & 1{,}685 \\
      \rowwhite Entertainment             &   537   \\
      \rowwhite Mentions                  &   425   \\
      \rowwhite Politics                  &   236   \\
      \rowwhite Companies                 &   108   \\
      \rowwhite Economics                 &    67   \\
      \rowwhite Elections                 &    64   \\
      \rowwhite Science and Technology    &    39   \\
      \rowwhite World                     &    23   \\
      \rowwhite Social                    &    12   \\
      \rowwhite Health                    &     8   \\
      \rowwhite Transportation            &     3   \\
      \rowwhite Education                 &     1   \\
      \midrule
      \rowwhite \textbf{Total}            & 3{,}208 \\
      \bottomrule
    \end{tabular}
  \end{minipage}
\end{table}

\section{Results for more input LLMs}
\label{app:more_results}
\begin{figure}[h]
  \begin{center}
    \centerline{\includegraphics[width=0.65\columnwidth]{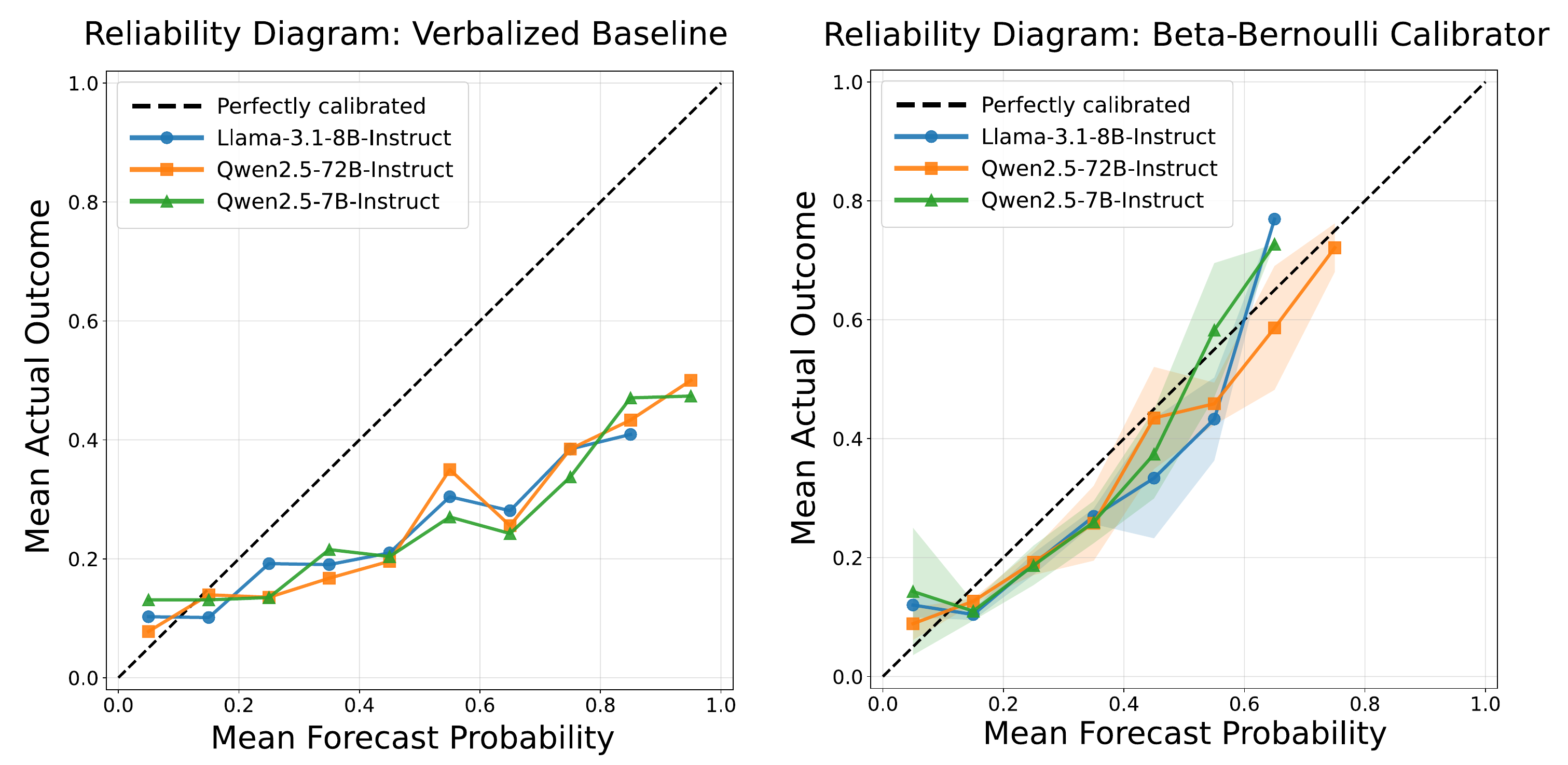}}
    \caption{
      Reliability Diagram for additional input LLMs. Verbalized forecasts exhibit overconfidence (left), and BBC improves calibration (right).
    }
    \label{fig:reliability_diagram_2}
  \end{center}
  \vskip -0.3in
\end{figure}

\begin{figure}[h]
  \begin{center}
    \centerline{\includegraphics[width=0.85\columnwidth]{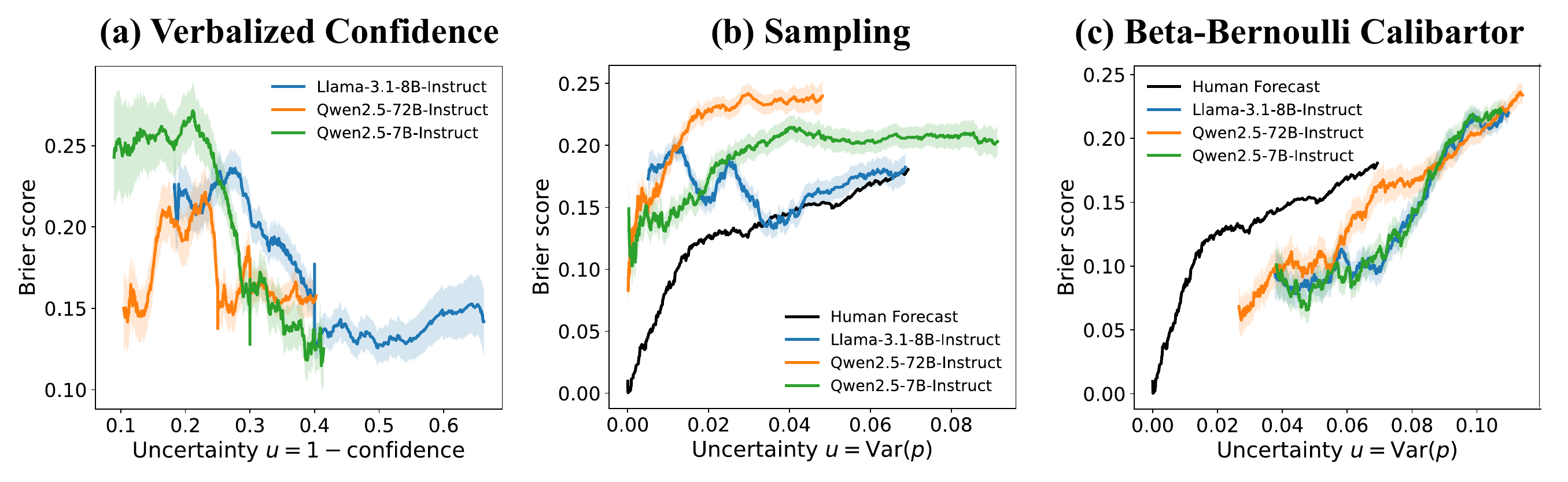}}
    \caption{The plot of Brier score against ranked epistemic uncertainty, smoothed with a window of 300. The uncertainty is defined as (a) 1 - verbalized confidence, (b) Sampling-based variance, and (c) BBC variance. The observation aligns with the discussion in Section~\ref{sec:analysis_u}.}
    \label{fig:u_vs_e_2}
    \vskip -0.3in
  \end{center}
  
\end{figure}

We report additional test results for the input LLMs Qwen2.5-72B-Instruct \citep{qwen2025qwen25technicalreport}, Qwen2.5-7B-Instruct \citep{qwen2025qwen25technicalreport}, and Llama-3.1-8B-Instruct \citep{grattafiori2024llama} in Table~\ref{tab:app_results}, Figure \ref{fig:reliability_diagram_2}, and Figure~\ref{fig:u_vs_e_2}. We notice consistent trends with our main findings, with BBC generally providing better forecasts than the baseline methods.

\begin{table*}[h]
  \caption{Test performance across additional input LLMs and baseline methods. Best results are bolded, and second-best results are underlined. KL is the KL divergence between the predicted distribution and the human forecast distribution on the test set.}
  \label{tab:app_results}
  \begin{center}
    \small
    \setlength{\tabcolsep}{3pt}
    \renewcommand{\arraystretch}{1}
    \begin{tabular}{@{}l rr rr rr rr rr@{}}
      \toprule
      \tableheader &
      \multicolumn{2}{c}{\tableheader Brier$\downarrow$} &
      \multicolumn{2}{c}{\tableheader Accuracy$\uparrow$} &
      \multicolumn{2}{c}{\tableheader AUC$\uparrow$} &
      \multicolumn{2}{c}{\tableheader ECE$\downarrow$} &
      \multicolumn{2}{c}{\tableheader KL$\downarrow$} \\
      \cmidrule(lr){2-3}\cmidrule(lr){4-5}\cmidrule(lr){6-7}\cmidrule(lr){8-9}\cmidrule(lr){10-11}
      \tableheader Input LLM / Method &
      \multicolumn{1}{c}{\tableheader mean} & \multicolumn{1}{c}{\tableheader std} &
      \multicolumn{1}{c}{\tableheader mean} & \multicolumn{1}{c}{\tableheader std} &
      \multicolumn{1}{c}{\tableheader mean} & \multicolumn{1}{c}{\tableheader std} &
      \multicolumn{1}{c}{\tableheader mean} & \multicolumn{1}{c}{\tableheader std} &
      \multicolumn{1}{c}{\tableheader mean} & \multicolumn{1}{c}{\tableheader std} \\
      \midrule

      \textbf{Qwen2.5-72B-Instruct} \\
      \rowwhite \hspace*{1.2em}Verbalized            & 0.174 &  & 0.764 &  & 0.655 &  & 0.144 &  &  &  \\
      \rowwhite \hspace*{1.2em}Ensemble ($n=10$)     & 0.165 &  & 0.766 &  & \secondbest{0.676} &  & 0.138 &  &  &  \\
      \rowwhite \hspace*{1.2em}Platt Scaling         & 0.138 &  & \secondbest{0.825} &  & 0.655 &  & 0.049 &  &  &  \\
      \rowwhite \hspace*{1.2em}Isotonic Regression   & 0.138 &  & \secondbest{0.825} &  & 0.655 &  & 0.044 &  &  &  \\
      \rowwhite \hspace*{1.2em}$P(\mathrm{True})$    & 0.261 &  & 0.736 &  & 0.633 &  & 0.262 &  &  &  \\
      \rowwhite \hspace*{1.2em}BBC (binary only)     & \secondbest{0.135} & (0.003) & \secondbest{0.825} & (0.007) & 0.670 & (0.005) & \secondbest{0.042} & (0.025) & 11.247 & (0.535) \\
      \rowblue  \hspace*{1.2em}BBC (binary+human)    & \best{0.133} & (0.003) & \best{0.829} & (0.002) & \best{0.683} & (0.009) & \best{0.035} & (0.017) & \best{9.160} & (0.556) \\
      \midrule

      \textbf{Qwen2.5-7B-Instruct} \\
      \rowwhite \hspace*{1.2em}Verbalized            & 0.170 &  & 0.778 &  & 0.621 &  & 0.124 &  &  &  \\
      \rowwhite \hspace*{1.2em}Ensemble ($n=10$)     & 0.159 &  & 0.797 &  & 0.646 &  & 0.121 &  &  &  \\
      \rowwhite \hspace*{1.2em}Platt Scaling         & 0.145 &  & 0.829 &  & 0.621 &  & 0.085 &  &  &  \\
      \rowwhite \hspace*{1.2em}Isotonic Regression   & 0.140 &  & 0.826 &  & 0.624 &  & \secondbest{0.064} &  &  &  \\
      \rowwhite \hspace*{1.2em}$P(\mathrm{True})$    & 0.166 &  & 0.827 &  & 0.568 &  & 0.165 &  &  &  \\
      \rowwhite \hspace*{1.2em}BBC (binary only)     & \secondbest{0.138} & (0.004) & \secondbest{0.830} & (0.004) & \secondbest{0.668} & (0.012) & 0.067 & (0.019) & 12.227 & (0.587) \\
      \rowblue  \hspace*{1.2em}BBC (binary+human)    & \best{0.135} & (0.002) & \best{0.831} & (0.007) & \best{0.676} & (0.006) & \best{0.054} & (0.018) & \best{9.677} & (0.385) \\
      \midrule

      \textbf{Llama-3.1-8B-Instruct} \\
      \rowwhite \hspace*{1.2em}Verbalized            & 0.169 &  & 0.771 &  & 0.639 &  & 0.140 &  &  &  \\
      \rowwhite \hspace*{1.2em}Ensemble ($n=10$)     & 0.165 &  & 0.766 &  & 0.662 &  & 0.153 &  &  &  \\
      \rowwhite \hspace*{1.2em}Platt Scaling         & 0.144 &  & \secondbest{0.827} &  & 0.639 &  & 0.069 &  &  &  \\
      \rowwhite \hspace*{1.2em}Isotonic Regression   & 0.143 &  & 0.823 &  & 0.637 &  & 0.060 &  &  &  \\
      \rowwhite \hspace*{1.2em}$P(\mathrm{True})$    & 0.217 &  & 0.768 &  & 0.623 &  & 0.216 &  &  &  \\
      \rowwhite \hspace*{1.2em}BBC (binary only)     & \secondbest{0.138} & (0.002) & 0.819 & (0.004) & \secondbest{0.669} & (0.004) & \best{0.055} & (0.012) & 11.703 & (0.336) \\
      \rowblue  \hspace*{1.2em}BBC (binary+human)    & \best{0.136} & (0.003) & \best{0.828} & (0.003) & \best{0.673} & (0.006) & \secondbest{0.058} & (0.015) & \best{9.815} & (0.281) \\
      \bottomrule
    \end{tabular}
  \end{center}
  \vskip -0.1in
\end{table*}

\section{Additional ablation studies}
\subsection{Ablation: number of mixture components $K$}
\label{app:ablation_K}

In our main experiments, we model $p$ as a mixture of $K=5$ Beta distributions. Here we study the effect of varying the number of mixture components $K \in \{1, 3, 5, 7, 10\}$. As shown in Table~\ref{tab:ablation_K}, $K=1$ runs notably underperform mixtures, while performance across $K=3$ to $K=10$ remains roughly stable.

\begin{table*}[t]
  \caption{Ablation on the number of mixture components $K$ in BBC.}
  \label{tab:ablation_K}
  \begin{center}
    \footnotesize
    \setlength{\tabcolsep}{3pt}
    \renewcommand{\arraystretch}{1}
    \begin{tabular}{@{}l rr rr rr rr@{}}
      \toprule
      \tableheader &
      \multicolumn{2}{c}{\tableheader Brier$\downarrow$} &
      \multicolumn{2}{c}{\tableheader Accuracy$\uparrow$} &
      \multicolumn{2}{c}{\tableheader AUC$\uparrow$} &
      \multicolumn{2}{c}{\tableheader ECE$\downarrow$} \\
      \cmidrule(lr){2-3}\cmidrule(lr){4-5}\cmidrule(lr){6-7}\cmidrule(lr){8-9}
      \tableheader Input LLM / $K$ &
      \multicolumn{1}{c}{\tableheader mean} & \multicolumn{1}{c}{\tableheader std} &
      \multicolumn{1}{c}{\tableheader mean} & \multicolumn{1}{c}{\tableheader std} &
      \multicolumn{1}{c}{\tableheader mean} & \multicolumn{1}{c}{\tableheader std} &
      \multicolumn{1}{c}{\tableheader mean} & \multicolumn{1}{c}{\tableheader std} \\
      \midrule
      \textbf{Claude-Sonnet-4} \\
      \rowwhite \hspace*{1.2em}$K=1$  & 0.129 & (0.002) & 0.831 & (0.005) & 0.737 & (0.002) & 0.053 & (0.018) \\
      \rowwhite \hspace*{1.2em}$K=3$  & \best{0.126} & (0.000) & \best{0.836} & (0.003) & \secondbest{0.742} & (0.008) & 0.038 & (0.002) \\
      \rowwhite  \hspace*{1.2em}$K=5$  & \best{0.126} & (0.001) & \best{0.836} & (0.003) & \best{0.744} & (0.005) & 0.036 & (0.007) \\
      \rowwhite \hspace*{1.2em}$K=7$  & \best{0.126} & (0.001) & \secondbest{0.835} & (0.003) & 0.740 & (0.002) & \secondbest{0.035} & (0.003) \\
      \rowwhite \hspace*{1.2em}$K=10$ & \best{0.126} & (0.002) & 0.834 & (0.004) & 0.737 & (0.005) & \best{0.034} & (0.004) \\
      \midrule
      \textbf{Llama-3.3-70B-Instruct} \\
      \rowwhite \hspace*{1.2em}$K=1$  & 0.145 & (0.003) & \secondbest{0.829} & (0.004) & 0.665 & (0.004) & 0.096 & (0.017) \\
      \rowwhite \hspace*{1.2em}$K=3$  & \secondbest{0.137} & (0.002) & 0.825 & (0.007) & \secondbest{0.677} & (0.009) & 0.065 & (0.010) \\
      \rowwhite  \hspace*{1.2em}$K=5$  & \best{0.136} & (0.001) & \best{0.830} & (0.001) & \best{0.679} & (0.012) & 0.060 & (0.005) \\
      \rowwhite \hspace*{1.2em}$K=7$  & 0.138 & (0.003) & 0.818 & (0.011) & 0.671 & (0.004) & \secondbest{0.052} & (0.024) \\
      \rowwhite \hspace*{1.2em}$K=10$ & \best{0.136} & (0.001) & 0.823 & (0.001) & 0.675 & (0.001) & \best{0.047} & (0.009) \\
      \bottomrule
    \end{tabular}
  \end{center}
\end{table*}

\subsection{Ablation: loss coefficients}
\label{app:ablation_loss_coef}

Given the training objective $\mathcal{L}_{\text{total}} = \lambda_{\text{binary}} \sum_i \mathcal{L}_{\text{binary},i} + \lambda_{\text{human}} \sum_i \mathcal{L}_{\text{human},i}$, we ablate the loss coefficients as shown in Table~\ref{tab:ablation_loss_coef}. Training on the human loss only ($\lambda_{\text{binary}}=0, \lambda_{\text{human}}=1$) achieves comparable performance to the binary+human training setup, validating that human forecasts are a useful supervision signal on their own. Compared to binary-only training, adding human supervision consistently improves performance with lower Brier score and higher AUC, and results remain relatively stable across a broad range of coefficients.

\begin{table*}[!t]
  \caption{Sensitivity to loss coefficients $\lambda_{\text{binary}}$ and $\lambda_{\text{human}}$. Adding human supervision consistently improves over binary-only training across a broad range of coefficients, and human-only training ($\lambda_{\text{binary}}=0$) is competitive with binary+human training, confirming that human forecasts are a useful supervision signal on their own.}
  \label{tab:ablation_loss_coef}
  \begin{center}
    \footnotesize
    \setlength{\tabcolsep}{3pt}
    \renewcommand{\arraystretch}{1}
    \begin{tabular}{@{}cc rr rr rr rr@{}}
      \toprule
      \tableheader & \tableheader &
      \multicolumn{2}{c}{\tableheader Brier$\downarrow$} &
      \multicolumn{2}{c}{\tableheader Accuracy$\uparrow$} &
      \multicolumn{2}{c}{\tableheader AUC$\uparrow$} &
      \multicolumn{2}{c}{\tableheader ECE$\downarrow$} \\
      \cmidrule(lr){3-4}\cmidrule(lr){5-6}\cmidrule(lr){7-8}\cmidrule(lr){9-10}
      \tableheader $\lambda_{\text{binary}}$ & \tableheader $\lambda_{\text{human}}$ &
      \multicolumn{1}{c}{\tableheader mean} & \multicolumn{1}{c}{\tableheader std} &
      \multicolumn{1}{c}{\tableheader mean} & \multicolumn{1}{c}{\tableheader std} &
      \multicolumn{1}{c}{\tableheader mean} & \multicolumn{1}{c}{\tableheader std} &
      \multicolumn{1}{c}{\tableheader mean} & \multicolumn{1}{c}{\tableheader std} \\
      \midrule
      \multicolumn{10}{l}{\textbf{Claude-Sonnet-4}} \\
      \rowwhite 1 & 0.0  & 0.127 & (0.002) & 0.834 & (0.003) & 0.729 & (0.011) & \secondbest{0.030} & (0.010) \\
      \rowwhite 1 & 0.1  & \best{0.125} & (0.001) & \secondbest{0.835} & (0.002) & 0.739 & (0.010) & \best{0.026} & (0.005) \\
      \rowwhite 1 & 0.5  & 0.127 & (0.001) & \secondbest{0.835} & (0.001) & 0.736 & (0.007) & 0.036 & (0.005) \\
      \rowwhite  1 & 1.0  & \secondbest{0.126} & (0.001) & \best{0.836} & (0.003) & \secondbest{0.744} & (0.005) & 0.036 & (0.007) \\
      \rowwhite 1 & 5.0  & \secondbest{0.126} & (0.002) & 0.834 & (0.001) & 0.743 & (0.017) & 0.034 & (0.003) \\
      \rowwhite 1 & 10.0 & \best{0.125} & (0.000) & \best{0.836} & (0.005) & \best{0.745} & (0.014) & 0.031 & (0.007) \\
      \rowwhite 0 & 1.0  & \best{0.125} & (0.000) & \best{0.836} & (0.005) & \secondbest{0.744} & (0.015) & 0.032 & (0.011) \\
      \midrule
      \multicolumn{10}{l}{\textbf{Llama-3.3-70B-Instruct}} \\
      \rowwhite 1 & 0.0  & 0.138 & (0.004) & 0.815 & (0.015) & 0.667 & (0.011) & 0.049 & (0.015) \\
      \rowwhite 1 & 0.1  & \best{0.135} & (0.001) & 0.824 & (0.003) & 0.671 & (0.003) & \best{0.039} & (0.011) \\
      \rowwhite 1 & 0.5  & \secondbest{0.136} & (0.002) & \secondbest{0.827} & (0.007) & 0.673 & (0.012) & \secondbest{0.047} & (0.008) \\
      \rowwhite  1 & 1.0  & \secondbest{0.136} & (0.001) & \best{0.830} & (0.001) & 0.679 & (0.012) & 0.060 & (0.005) \\
      \rowwhite 1 & 5.0  & 0.137 & (0.003) & 0.826 & (0.012) & \secondbest{0.683} & (0.005) & 0.067 & (0.011) \\
      \rowwhite 1 & 10.0 & 0.137 & (0.004) & 0.825 & (0.012) & \best{0.684} & (0.003) & 0.062 & (0.016) \\
      \rowwhite 0 & 1.0  & 0.137 & (0.005) & 0.824 & (0.010) & \best{0.684} & (0.002) & 0.063 & (0.021) \\
      \bottomrule
    \end{tabular}
  \end{center}
   \vskip -0.2in
\end{table*}

\subsection{Robustness to sparse/biased human forecasts}
\label{app:robustness_human}

A natural concern with using human forecasts as supervision is whether BBC remains effective when they are sparse or biased. We note that human forecasts in our training set are well-calibrated and significantly outperform all LLMs (human: Brier $=0.085$, AUC $=0.945$ vs. model average: Brier $=0.196$, AUC $=0.704$), validating their quality as supervision. The corruption experiments below therefore serve as stress tests rather than reflections of realistic settings.

\subsubsection{Sparsity}

To simulate sparse human signals, we retain $x\%$ of the forecasts per question during training. Table~\ref{tab:robustness_sparsity} shows that BBC performance is largely robust to forecast sparsity: even with only 10\% of forecasts retained, BBC still shows gains in AUC, indicating that even sparse human forecasts provide useful distributional signals. As more human signals become available, we observe improvements across both input LLMs.

\begin{table*}[!t]
  \caption{Robustness to sparse human forecasts.}
  \label{tab:robustness_sparsity}
  \begin{center}
    \footnotesize
    \setlength{\tabcolsep}{3pt}
    \renewcommand{\arraystretch}{1}
    \begin{tabular}{@{}l rr rr rr rr@{}}
      \toprule
      \tableheader &
      \multicolumn{2}{c}{\tableheader Brier$\downarrow$} &
      \multicolumn{2}{c}{\tableheader Accuracy$\uparrow$} &
      \multicolumn{2}{c}{\tableheader AUC$\uparrow$} &
      \multicolumn{2}{c}{\tableheader ECE$\downarrow$} \\
      \cmidrule(lr){2-3}\cmidrule(lr){4-5}\cmidrule(lr){6-7}\cmidrule(lr){8-9}
      \tableheader Input LLM / \% Retained &
      \multicolumn{1}{c}{\tableheader mean} & \multicolumn{1}{c}{\tableheader std} &
      \multicolumn{1}{c}{\tableheader mean} & \multicolumn{1}{c}{\tableheader std} &
      \multicolumn{1}{c}{\tableheader mean} & \multicolumn{1}{c}{\tableheader std} &
      \multicolumn{1}{c}{\tableheader mean} & \multicolumn{1}{c}{\tableheader std} \\
      \midrule
      \textbf{Claude-Sonnet-4} \\
      \rowwhite \hspace*{1.2em}0\% (binary only) & \secondbest{0.127} & (0.002) & \secondbest{0.834} & (0.003) & 0.729 & (0.011) & \best{0.030} & (0.010) \\
      \rowwhite \hspace*{1.2em}10\%   & 0.129 & (0.001) & 0.831 & (0.001) & 0.738 & (0.011) & 0.061 & (0.006) \\
      \rowwhite \hspace*{1.2em}25\%   & \secondbest{0.127} & (0.001) & 0.832 & (0.006) & \secondbest{0.743} & (0.007) & 0.048 & (0.003) \\
      \rowwhite \hspace*{1.2em}50\%   & 0.128 & (0.001) & 0.831 & (0.007) & 0.739 & (0.011) & 0.047 & (0.003) \\
      \rowwhite  \hspace*{1.2em}100\%  & \best{0.126} & (0.001) & \best{0.836} & (0.003) & \best{0.744} & (0.005) & \secondbest{0.036} & (0.007) \\
      \midrule
      \textbf{Llama-3.3-70B-Instruct} \\
      \rowwhite \hspace*{1.2em}0\% (binary only) & 0.138 & (0.004) & 0.815 & (0.015) & 0.667 & (0.011) & \best{0.049} & (0.015) \\
      \rowwhite \hspace*{1.2em}10\%   & \secondbest{0.137} & (0.003) & \secondbest{0.829} & (0.009) & \secondbest{0.676} & (0.005) & 0.071 & (0.012) \\
      \rowwhite \hspace*{1.2em}25\%   & \secondbest{0.137} & (0.003) & 0.823 & (0.013) & \best{0.679} & (0.009) & 0.063 & (0.010) \\
      \rowwhite \hspace*{1.2em}50\%   & \best{0.136} & (0.000) & 0.828 & (0.005) & 0.674 & (0.009) & 0.067 & (0.005) \\
      \rowwhite  \hspace*{1.2em}100\%  & \best{0.136} & (0.001) & \best{0.830} & (0.001) & \best{0.679} & (0.012) & \secondbest{0.060} & (0.005) \\
      \bottomrule
    \end{tabular}
  \end{center}
   \vskip -0.1in
\end{table*}

\subsubsection{Bias}

We further test BBC's robustness under three types of synthetic corruption applied to the human forecasts during training:
\begin{itemize}[leftmargin=*]
    \item \textbf{Noise:} replacing a fraction of forecasts with $\text{Uniform}(0,1)$ draws;
    \item \textbf{Directional shift ($\gamma$):} scaling each forecast as $q' = 0.5 + \gamma(q - 0.5)$, where $\gamma<1$ pulls forecasts toward 0.5 (underconfident) and $\gamma>1$ pushes them toward the extremes (overconfident);
    \item \textbf{Additive shift ($\delta$):} shifting all forecasts by a constant $q' = q + \delta$, where $\delta>0$ is optimistic and $\delta<0$ is pessimistic.
\end{itemize}

Table~\ref{tab:robustness_corruption} shows that systematic underconfidence and noise are the most damaging, as both push the predicted Beta distribution toward flat and uninformative shapes. AUC remains relatively robust across all corruption types, since systematic bias preserves the relative ordering among predictions. Interestingly, mild overconfidence ($\gamma=1.5$) and negative additive shift ($\delta=-0.1$) actually sometimes improve performance over the uncorrupted baseline. This reflects the class imbalance in our test set: 83\% of events resolve to ``No'', so corruptions that push forecasts toward 0 (negative shift) or sharpen them away from 0.5 (overconfidence) tend to align with the majority outcome.

\begin{table*}[!t]
  \caption{Robustness to corrupted human forecasts.}
  \label{tab:robustness_corruption}
  \begin{center}
    \footnotesize
    \setlength{\tabcolsep}{3pt}
    \renewcommand{\arraystretch}{1}
    \begin{tabular}{@{}l c rr rr rr rr@{}}
      \toprule
      \tableheader & \tableheader &
      \multicolumn{2}{c}{\tableheader Brier$\downarrow$} &
      \multicolumn{2}{c}{\tableheader Accuracy$\uparrow$} &
      \multicolumn{2}{c}{\tableheader AUC$\uparrow$} &
      \multicolumn{2}{c}{\tableheader ECE$\downarrow$} \\
      \cmidrule(lr){3-4}\cmidrule(lr){5-6}\cmidrule(lr){7-8}\cmidrule(lr){9-10}
      \tableheader Corruption & \tableheader Parameter &
      \multicolumn{1}{c}{\tableheader mean} & \multicolumn{1}{c}{\tableheader std} &
      \multicolumn{1}{c}{\tableheader mean} & \multicolumn{1}{c}{\tableheader std} &
      \multicolumn{1}{c}{\tableheader mean} & \multicolumn{1}{c}{\tableheader std} &
      \multicolumn{1}{c}{\tableheader mean} & \multicolumn{1}{c}{\tableheader std} \\
      \midrule
      \textbf{Claude-Sonnet-4} \\
      \rowwhite \hspace*{1.2em}Binary only         & ---     & 0.127 & (0.002) & 0.834 & (0.003) & 0.729 & (0.011) & \best{0.030} & (0.010) \\
      \rowwhite \hspace*{1.2em}Noise               & 25\%    & 0.131 & (0.003) & \secondbest{0.836} & (0.005) & 0.740 & (0.008) & 0.084 & (0.013) \\
      \rowwhite \hspace*{1.2em}Noise               & 50\%    & 0.144 & (0.003) & 0.833 & (0.008) & 0.728 & (0.007) & 0.138 & (0.012) \\
      \rowwhite \hspace*{1.2em}Underconfident ($\gamma$) & 0.5  & 0.157 & (0.001) & \best{0.838} & (0.003) & 0.733 & (0.002) & 0.174 & (0.001) \\
      \rowwhite \hspace*{1.2em}Overconfident ($\gamma$)  & 1.5  & \best{0.125} & (0.002) & 0.833 & (0.004) & \best{0.745} & (0.008) & \secondbest{0.032} & (0.002) \\
      \rowwhite \hspace*{1.2em}Negative shift ($\delta$) & $-0.1$ & \best{0.125} & (0.002) & 0.835 & (0.001) & 0.740 & (0.005) & \best{0.030} & (0.003) \\
      \rowwhite \hspace*{1.2em}Positive shift ($\delta$) & $+0.1$ & 0.137 & (0.006) & 0.829 & (0.008) & 0.739 & (0.002) & 0.109 & (0.028) \\
      \rowwhite  \hspace*{1.2em}No corruption       & ---     & \secondbest{0.126} & (0.001) & \secondbest{0.836} & (0.003) & \secondbest{0.744} & (0.005) & 0.036 & (0.007) \\
      \midrule
      \textbf{Llama-3.3-70B-Instruct} \\
      \rowwhite \hspace*{1.2em}Binary only         & ---     & 0.138 & (0.004) & 0.815 & (0.015) & 0.667 & (0.011) & 0.049 & (0.015) \\
      \rowwhite \hspace*{1.2em}Noise               & 25\%    & 0.144 & (0.005) & 0.826 & (0.008) & 0.672 & (0.015) & 0.107 & (0.017) \\
      \rowwhite \hspace*{1.2em}Noise               & 50\%    & 0.155 & (0.005) & 0.824 & (0.010) & 0.673 & (0.005) & 0.150 & (0.017) \\
      \rowwhite \hspace*{1.2em}Underconfident ($\gamma$) & 0.5  & 0.170 & (0.003) & 0.821 & (0.011) & 0.672 & (0.008) & 0.190 & (0.010) \\
      \rowwhite \hspace*{1.2em}Overconfident ($\gamma$)  & 1.5  & \best{0.134} & (0.002) & 0.828 & (0.003) & \best{0.684} & (0.005) & \best{0.030} & (0.003) \\
      \rowwhite \hspace*{1.2em}Negative shift ($\delta$) & $-0.1$ & \best{0.134} & (0.001) & \secondbest{0.829} & (0.005) & 0.678 & (0.003) & \secondbest{0.035} & (0.019) \\
      \rowwhite \hspace*{1.2em}Positive shift ($\delta$) & $+0.1$ & 0.148 & (0.003) & 0.821 & (0.007) & 0.674 & (0.001) & 0.125 & (0.010) \\
      \rowwhite  \hspace*{1.2em}No corruption       & ---     & \secondbest{0.136} & (0.001) & \best{0.830} & (0.001) & \secondbest{0.679} & (0.012) & 0.060 & (0.005) \\
      \bottomrule
    \end{tabular}
  \end{center}
\end{table*}

\section{Prompts}
\label{app:prompts}
The verbalized probabilistic forecasts are elicited using the prompt in Figure \ref{fig:prompt_verb}. We use greedy decoding with temperature = 0, except for the ensemble method, where we use temperature = 1. Figure \ref{fig:prompt_verb_u} shows the prompt used to obtain verbalized confidence for the epistemic uncertainty analysis.

\begin{figure}[h]
    \centering
\begin{mdframed}
\small
You're an expert in forecasting events. Make a prediction of the probability that the question will be resolved as true. You MUST give a probability estimate between 0 and 1 UNDER ALL CIRCUMSTANCES. If for some reason you can’t answer, pick the base rate, but return a number between 0 and 1.\\
\\
To support your reasoning, recall relevant recent events, facts, or widely known information. Ensure your rationale is well-grounded and coherent.  \\
\\
Once you have completed your reasoning, output your answer as a number between 0 and 1.\\
\\
Question: \{\}\\
Resolution Criteria: \{\}\\
\\
Today’s date: \{\}\\
Question close date: \{\}\\
\\
Please follow the output format:\\
{[Rationale:]} xxx \\
{[Answer:]} a number between 0 and 1\\
\end{mdframed}
 \caption{Prompt for obtaining verbalized forecasts from the input LLMs.}
 \label{fig:prompt_verb}
\end{figure}

\begin{figure}[h]
    \centering
\begin{mdframed}
\small
You're an expert in forecasting events. Make a prediction of the probability that the question will be resolved as true. You MUST give a probability estimate between 0 and 1 UNDER ALL CIRCUMSTANCES. If for some reason you can’t answer, pick the base rate, but return a number between 0 and 1.\\
\\
To support your reasoning, recall relevant recent events, facts, or widely known information. Ensure your rationale is well-grounded and coherent.\\
\\
Once you have completed your reasoning, output your answer as a number between 0 and 1.\\
\\
After you give your probability, also report how confident you are in that probability on a scale from 0 to 1 (0 = no confidence, 1 = extremely confident).\\
\\
Question: \{\}\\
Resolution Criteria: \{\}\\
\\
Today’s date: \{\}\\
Question close date: \{\}\\
\\
Please follow the output format:\\
{[Rationale:]} xxx\\
{[Answer:]} a number between 0 and 1\\
{[Confidence:]} a number between 0 and 1\\
\end{mdframed}
 \caption{Prompt for obtaining verbalized forecasts together with verbalized confidence in that forecast.}
 \label{fig:prompt_verb_u}
\end{figure}

\end{document}